\documentclass[journal]{IEEEtran}
\usepackage{graphicx}
\usepackage{amsmath}
\usepackage{amssymb}
\usepackage{color}
\usepackage{bbding}
\usepackage{bbm}
\usepackage{booktabs}
\usepackage{threeparttable}
\usepackage{subfigure}
\usepackage[ruled]{algorithm2e}

\begin{document}


\title{Hierarchical Knowledge Guided Learning for Real-world Retinal Diseases Recognition}

\author{Lie~Ju,
        Zhen~Yu,
        Lin~Wang,
        Xin~Zhao,
        Xin~Wang,
        Paul~Bonnington,
        and~Zongyuan Ge
\thanks{(Corresponding author: Zongyuan Ge)}
\thanks{Lie Ju, Zhen Yu, Paul Bonnington, and Zongyuan Ge are with Monash University, Melbourne, VIC 3800 Australia; Lie Ju, Zhen Yu, and Zongyuan Ge are also with the Monash-Airdoc joint research group, Monash University, Melbourne, VIC 3800 Australia (E-mail: julie334600@gmail.com, zhen.yu@monash.edu, paul.bonnington@monash.edu, zongyuan.ge@monash.edu).}
\thanks{Lin Wang is with the Monash-Airdoc joint research group, Monash University, Melbourne, VIC 3800 Australia (E-mail: wanglin.mailbox@gmail.com).}
\thanks{Xin Wang and Xin Zhao are with Airdoc 100089, Beijing, China (E-mail: {wangxin, zhaoxin}@airdoc.com).}
}

\markboth{Journal of \LaTeX\ Class Files,~Vol.~14, No.~8, August~2015}%
{Shell \MakeLowercase{\textit{et al.}}: Bare Demo of IEEEtran.cls for IEEE Journals}


\maketitle

\begin{abstract}
In the real world, medical datasets often exhibit a long-tailed data distribution (i.e., a few classes occupy the majority of the data, while most classes have only a limited number of samples), which results in a challenging \textit{long-tailed} learning scenario. 
Some recently published datasets in ophthalmology AI consist of more than 40 kinds of retinal diseases with complex abnormalities and variable morbidity. Nevertheless, more than 30 conditions are rarely seen in global patient cohorts. 
From a modeling perspective, most deep learning models trained on these datasets may lack the ability to generalize to rare diseases where only a few available samples are presented for training. 
In addition, there may be more than one disease for the presence of the retina, resulting in a challenging label co-occurrence scenario, also known as \textit{multi-label}, which can cause problems when some re-sampling strategies are applied during training. 
To address the above two major challenges, this paper presents a novel method that enables the deep neural network to learn from a long-tailed fundus database for various retinal disease recognition. 
Firstly, we exploit the prior knowledge in ophthalmology to improve the feature representation using a hierarchy-aware pre-training. 
Secondly, we adopt an instance-wise class-balanced sampling strategy to address the label co-occurrence issue under the long-tailed medical dataset scenario. 
Thirdly, we introduce a novel hybrid knowledge distillation to train a less biased representation and classifier. 
We conducted extensive experiments on four databases, including two public datasets and two in-house databases with more than one million fundus images. The experimental results demonstrate the superiority of our proposed methods with recognition accuracy outperforming the state-of-the-art competitors, especially for these rare diseases.

\end{abstract}


\section{Introduction}

Retinal diseases such as diabetic retinopathy (DR) and glaucoma are the leading causes of blindness~\cite{Klein2007Overview,gulshan2016development}. These diseases usually have mild manifestations in the early stages, but would cause irreversible damage to the optic nerves in the later stages~\cite{Kocur2002Visual}. Triage screening with a fundus camera or optical coherence tomography (OCT) instrument is essential for early detection of these retinal diseases. However, manual fundus screening requires ophthalmologists to review and report fundus images, which is time-consuming. 

The automated screening of retinal diseases has long been recognized and has attracted much attention~\cite{Faust2012Algorithms,li2021applications}. Recent studies have demonstrated the success of deep learning-based models for screening retinal diseases such as diabetic retinopathy (DR) and glaucoma~\cite{gulshan2016development,fu2018disc}. Gargeya et al.~\cite{gargeya2017automated} used features extracted from CNN and metadata information fed into a decision tree model for binary classification. Ara{\'u}jo et al.~\cite{araujo2020dr} proposed a multi-instance learning and uncertainty-based framework for DR grading. For glaucoma detection, the optic disc-to-cup (OD/OC) ratio is always considered. Dos et al.~\cite{dos2018convolutional} designed a phylogenetic diversity index module for semantic feature extraction for glaucoma diagnosis. Fu et al.~\cite{fu2018disc} proposed a novel disc-aware ensemble network, which integrates the deep hierarchical context of the global fundus image and the local optic disc region. 
 
\begin{figure}[t]
	\includegraphics[width=8.5cm]{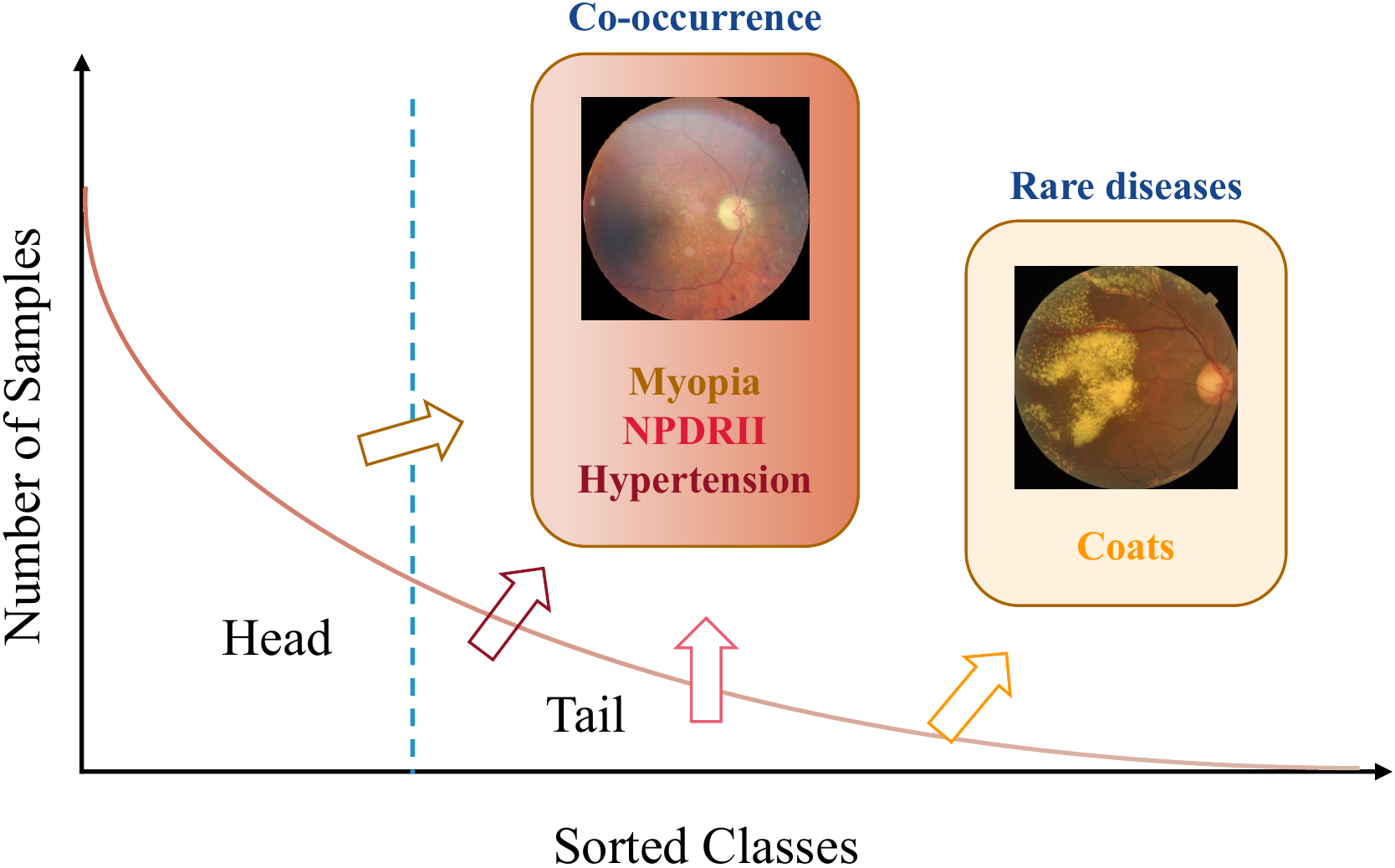}
	\centering
\caption{The retinal disease distribution of~\cite{ju2022improving} exhibits a prominent long-tailed characteristic. The co-occurrence of labels is also a common observation among the head, medium, and tail classes. } \label{fig_disease}
\end{figure}
Many existing works proved to be effective for detecting some specific diseases and lesions under a controlled test environment~\cite{gulshan2016development,fu2018disc,araujo2020dr}. However, the pathological changes in the fundus are diverse and heterogeneous in appearance, and comorbidity is difficult to define. For example, more than 40 types of retinal diseases have been categorized in modern ophthalmology research~\cite{steinmetz2021causes,ju2021relational}, and many of them, such as Coats' disease, retinoschisis, are rare diseases, resulting in a long-tail characteristic in terms of pathology distribution for modeling. From a machine learning perspective, those diseases and lesions that are underrepresented in the training set may not perform well in real clinical testing settings due to a lack of recognition of the corresponding pathologies. Furthermore, diagnosing the presence of multiple diseases is very critical in clinical practice, especially for those rare diseases that are easily ignored clinically.

Recently, the diagnosis of multiple retinal diseases, often defined as a multi-label classification task, using deep neural networks has attracted much attention. Wang et al.~\cite{wang2019retinal} developed a three-stream framework to detect 36 retinal lesions/diseases. Quellec et al.~\cite{quellec2020automatic} used few-shot learning to perform rare pathology detection on the OPHDIAT dataset~\cite{massin2008ophdiat}, which consists of 763,848 fundus images with 41 conditions. Ju et al.~\cite{ju2021relational} used prior knowledge (e.g., the region information) to improve the detection of 48 retinal diseases. A recently published dataset~\cite{ju2022improving} covers 53 classes of fundus diseases that may appear in clinical screening. As shown in Fig.~\ref{fig_disease}, this dataset has a typical long-tailed distribution and the ratio of head classes to tail classes exceeds 100. The problem of severe class imbalance is very common in many medical datasets. Although the above methods show promising results for the diagnosis of various retinal diseases, there are obvious limitations: (1) The human expert knowledge in the medical domain/ophthalmology, such as hierarchical information, is underrepresented. Leveraging such information helps to alleviate the training difficulty in a more orderly way~\cite{wang2019retinal,ju2021relational}.
(2) Some cases may have more than one retinal disease, leading to a label co-occurrence problem~\cite{wu2020distribution}. (3) Most existing works only train the model from a small database (less than one hundred thousand images), and therefore the generalization ability of the models in a real-world scenario is hardly guaranteed.

In this work, we propose a novel framework for real-world retinal disease recognition. Our proposed method is well applicable to learning from \textbf{long-tailed} data distributions. Our motivation is based on two main observations. First, incorporating the hierarchical information into the model would benefit the feature representation learning for those similar classes with shared features, which can come from majority and minority classes~\cite{dhall2020hierarchical}. Second, the model learning from the original imbalanced distribution obtains a better feature representation, i.e., convolutional layers, while the model learning from a rebalanced distribution shows a fairer and more unbiased classifier~\cite{zhou2020bbn,kang2019decoupling}, i.e., a fully connected layer.
Following these two observations, we first exploit the prior knowledge in retinal diseases, i.e., the hierarchical information for all categories organized from coarse to fine, to help the model learn better representations in a hierarchy-aware manner. Then, we use the regular instance-balanced sampling and a novel instance-wise class-balanced sampling to train two teacher models for the feature-balanced representation and less biased classifier, respectively. Finally, we distill the knowledge from the two teacher models into a unified student model.
Our main contributions are summarized below:

\begin{enumerate}
    
    \item We propose to improve the learning of the feature representation during the model training by injecting hierarchical information as prior knowledge. This method can learn a well-generalized feature representation from semantic information, which facilitates the recognition of similar diseases.
    
    \item We propose a novel hybrid knowledge distillation method to help the model learn from multiple teacher models simultaneously. This results in a feature-balanced representation and a less biased classifier learned by the student model.

    \item Experimental results on two public and two in-house datasets demonstrate that our methods can generalize well to various datasets and surpass the existing state-of-the-art methods.
    
    \item To the best of our knowledge, this is the first work to recognize more than 50 kinds of retinal diseases from the in-house database of more than \textbf{one million} fundus images. We collaborate with more than 10 experts in ophthalmology and release a carefully designed three-level hierarchical tree covering more than 100 retinal diseases. We hope that it can contribute to the research community to address the challenges.

\end{enumerate}



\section{Related Work}
\subsection{Retinal Diseases Recognition}

The computer-assisted diagnosis (CAD) methods for retinal diseases such as DR, Glaucoma and age-related macular degeneration (AMD) have been long recognized.  Gulshan et al.~\cite{gulshan2016development} and Gargeya et al.~\cite{gargeya2017automated} proposed to use a CNN for binary classification of with/without DR. Zhang et al.~\cite{zhang2019automated} presented an ensemble strategy to perform two-class and four-class classifications. Some methods~\cite{yang2017lesion,gondal2017weakly,foo2020multi} proposed to combine the segmentation results and grading results to provide the interpretability with the internal correlation between DR levels and lesions. Besides the OC/OD area-based method~\cite{fu2018disc} mentioned above, there are some previous works that generate evidence maps for glaucoma diagnosis. Zhao et al.~\cite{zhao2019direct} proposed a two-stage cascaded approach that obtains unsupervised feature representation of fundus image with a CNN and cup-disc-ratio value regression by random forest regressor. The automated diagnosis of AMD has also been studied~\cite{govindaiah2018deep,burlina2018use}. Ju et al.~\cite{ju2021synergic} leveraged the common features between DR and AMD, then improve both accuracy under a knowledge distillation and multi-task manner. 

Although some specific retinal diseases are fully studied, the condition of the fundus is complex. Most of the existing methods do not promise the robustness and ability to diagnose other diseases, especially rare ones. Kaggle EyePACS dataset~\cite{KagleDiabetic} consists of 35,126 training images graded into five DR stages. However, it is found that there are at least more than 30 kinds of retinal diseases that are mislabeled by the original annotators~\cite{ju2022improving}. Ignoring those out-of-distribution categories leads to a huge risk. Wang et al.~\cite{wang2019retinal} first used a multi-task framework to detect 36 kinds of retinal diseases, respectively. However, it requires extra annotations for the location of the optic disc and macula area. Quellec et al.~\cite{quellec2020automatic} also took the features among diseases into consideration and leveraged the few-shot learning technique to perform rare pathologies recognition. The existing methods are inspiring but show some limitations: those methods consider the prior knowledge of retinal diseases, such as the common features (lesions) and locations shared by some diseases, but still lack direct insights for leveraging some prior knowledge or cognitive laws.

\subsection{Long-tailed Classification}
We group existing methods for long-tailed learning into three main categories based on their main technical contributions, i.e., re-sampling, re-weighting, and feature sharing.

\textbf{Re-smapling }methods try to balance the distribution by over-sampling the minority-class samples or under-sampling the majority-class samples. Kang et al.~\cite{kang2019decoupling} proposed a two-stage training strategy to train the feature extractor and classifier from a uniform and re-sampling distribution, respectively. Zhou et al.~\cite{zhou2020bbn} proposed to design a two-branch network to achieve one-stage training with the same idea. Wang et al.~\cite{wang2019dynamic} presented a curriculum learning-based sampling scheduler. Most of these works focus on single-label datasets. For multi-label datasets, over-sampling of minority class will sometimes sample the majority class at the same time due to the label co-occurrence and therefore leads to a new imbalanced condition. To handle this challenge, Wu et al.~\cite{wu2020distribution} extended the re-sampling strategy into a multi-label scenario and introduced a regularization term to overcome the over-suppression from negative samples.

\textbf{Re-weighting }methods aim to design robust loss functions for learning from imbalanced data. Focal loss~\cite{lin2017focal} computes weight for each training sample and achieves Hard Samples Mining according to the prediction probabilities. The increase in the sample brings about a diminishing return on performance for those majority classes. Cui et al.~\cite{cui2019class} proposed to assign a new variable to increase the benefit from those effective samples. Cao et al.~\cite{cao2019learning} proposed an effective training strategy, which allows the model to learn an initial representation while avoiding some of the complications associated with re-weighting or re-sampling.

\textbf{Feature Sharing }methods aim to learn the general knowledge from the majority classes and then transfer it to the minority classes. Liu et al. \cite{liu2019large} proposed OLTR, which learns a set of dynamic meta embedding to transfer the visual information knowledge of the head to the tail category. Also, they designed a memory set that allows tail categories to utilize relevant head information by calculating the similarity. Xiang et al.~\cite{xiang2020learning} found that learning from a less-imbalanced subset suffers from little performance loss and proposed to train multiple shot-based teacher models, then guide the training of a unified student model. Li et al.~\cite{li2020overcoming} proposed balanced group softmax (BAGS) to improve long-tailed object detection by grouping the categories with similar numbers of training instances. BAGS implicitly modulates the head and tail classes are sufficiently trained, without requiring any extra sampling for the tail classes.


\section{Datasets}
\label{Datasets}

\subsection{Dataset Definition}
To address the challenge of long-tailed retinal diseases recognition, we evaluate the proposed methods on two in-house and two public datasets with different scales, including the number of samples, the number of classes, the imbalance ratio, and the degree of label co-occurrence, e.g., label cardinality. 

Here, we give the quantitative metrics for the main characteristics of long-tailed retinal disease recognition. The first is how imbalanced the original distribution is. Formally, for the original distribution $S = \{(x_{1},y_{1}), (x_{2},y_{2}), ..., (x_{n},y_{n})\}$ where each $x$ indicates a feature vector and each $y$ is an associated one-hot label. The indexes of \emph{k} kinds of categories are sorted from most to least with $N_{c_{1}} > N_{c_{2}} > ... > N_{c_{k}}$ where $N_{c_{*}}$ denotes the number of instances of the $*_{th}$ category. Thus, the imbalance ratio can be simply described as $\rho = \frac{N_{c_{1}}}{N_{c_{k}}}$. Moreover, as we claimed that there is label co-occurrence among the original long-tailed distribution, and most re-sampling strategies will lead to a new inner-class imbalance~\cite{wu2020distribution} (we have a detailed analysis of this in Sec.~\ref{sec. sampling}). Here, we introduce several useful indicators for measuring the label-occurrence~\cite{read2011classifier,zhang2013review}. The first is \emph{label cardinality} which indicates the average number of labels per example: $L_{Card}(S) = \frac{1}{n}\Sigma^{n}_{i=1}|y_{i}|$. The characteristics of the four datasets are shown in Table~\ref{table. dataset}.

\begin{table}[t]
\centering
\caption{The data statistics of four datasets.}
\small
\label{table. dataset}
\begin{tabular}{c|cccc}
\hline \hline
Dataset    & RFMiD  & ODIR   & Retina-100K & Retina-1M \\ \hline
Train      & 1920   & 7000   & 75714       & 839890    \\
Val        & 640    & 1000   & 9335        & 104981    \\
Test       & 640    & 2000   & 9477        & 104987    \\
Class      & 29     & 12     & 48          & 53        \\
$\rho$     & 80     & 309.8  & 828.56      & 78782.86  \\
$L_{Card}$ & 1.2864 & 1.1142 & 1.3439      & 1.5046    \\ \hline \hline

\end{tabular}
\end{table}

\subsection{Public Datasets}
\begin{table}[t]
\centering
\caption{The designed hierarchy of ODIR dataset.}
\begin{tabular}{l|l}
\hline\hline
Coarse                   & Fine                        \\ \hline
Normal                   & Normal                      \\
DR                       & NPDRI, NPDRII, \\ & NPDRIII, PDR \\
Cataract                 & Cataract                    \\
Glaucoma                 & Glaucoma                    \\
AMD                      & Dry AMD, Wet AMD            \\
Myopia                   & Pathological myopia         \\
Hypertensive retinopathy & Hypertensive retinopathy    \\
Others                   & Others                      \\ \hline \hline
\end{tabular}
\label{table_odir_dataset}
\end{table}

\subsubsection{RFMiD}
\label{sec. rfmid}
The RFMiD dataset~\cite{RFMiD} consists of 3200 fundus images captured using three different fundus cameras with 46 conditions annotated through adjudicated consensus of two senior retinal experts. The RFMiD dataset was originally divided into 1,920 / 640 / 640 images for training / validation / testing. We followed the setting in the RFMiD challenge, the diseases with more than 10 images belong to an independent class and all other disease categories are merged as “OTHER”. This finally constitutes 29 classes (normal + 28 diseases or lesions) for disease classification. Depending on the ratio of the number of samples in a category to the category with the largest number of samples, we assign it to one of the groups many, medium, or few. In the RFMiD dataset, given sorted 29 classes, we have 6 / 8 / 15 classes for groups \textit{many} / \textit{medium} / \textit{few}, respectively. Noted that the annotated classes in RFMiD are a blend of diseases and lesions. That is, a lesion may be present in more than one disease. To address this issue and to better utilize the proposed methods described in Sec.~\ref{Sec. hierarchy}, we collaborated with experts with rich experience in ophthalmology and build a hierarchy tree with four levels for the RFMiD dataset. For more details, please refer to our supplementary files.

\subsubsection{ODIR}
Ocular Disease Intelligent Recognition (ODIR)~\cite{ODIR} is a structured ophthalmic database of 5,000 patients with age, color fundus photographs of left and right eyes, and doctors' diagnostic keywords. Specifically, the ODIR dataset was originally divided into 8,000 / 1,000 / 2,000 images for training / off-site testing / on-site testing. Annotations were provided by trained human readers with quality control management. The classes of annotations can be divided into two levels: coarse and fine. There are 8 classes at the coarse level: Normal (N), Diabetes (D), Glaucoma (G), Cataract (C), Age-Related Macular Degeneration (A), Hypertension (H), Pathologic Myopia (M), and Other diseases/abnormalities (O), where one or more conditions are given on a patient-level diagnosis. The fine-level annotation is given on the left/right eye. In this work, we selected 12 fine classes to build the hierarchical tree, and the details are shown in Table~\ref{table_odir_dataset}. In the ODIR dataset, given sorted 12 classes, we have 3 / 6 / 3 classes for groups \textit{many} / \textit{medium} / \textit{few}, respectively. We directly exploit the coarse/fine annotation design for our proposed hierarchy-aware pre-training.

\subsection{In-house Datasets}
Our collected private datasets were acquired from private hospitals over the time span of 10 years. Some commonly used datasets are also included, such as ODIR and RFMiD. Each image was labeled or relabeled by 3 - 8 senior ophthalmologists. A sample is retained only if more than half of the ophthalmologists are in agreement with the disease label, or the sample will be assigned to re-labeled processing with the discussion of all ophthalmologists. In this study, more than one million samples are selected to form the two datasets, \textit{Retina-100K} and \textit{Retina-1M}, which consist of more than 50 kinds of retinal diseases. The Retina-100 K dataset is randomly sampled from the Retina-1M dataset, some rare diseases with no more than 10 instances are not sampled, resulting in the number of categories being only 48. It should be noted that some low-quality fundus images are potentially at risk of being misdiagnosed as cataracts. Therefore, a group of quality measurement categories is also added. In addition, we found that some non-fundus images could also be mistakenly predicted as retinal diseases during the inference stage. Similarly, we added a category for non-fundus images to reduce this risk. 

We would like to emphasize that this is the first study that attempts to train the DL-based model for retinal disease recognition from a database with more than \textbf{one million} samples. It can be found that the \emph{Retina} dataset exhibits an extreme long-tailed distribution with $\rho$ closing to 80,000. It is noticed that, with the increase in the number of samples, the samples of various categories are also enriched, and the label co-occurrence is reduced (e.g., 1.3439 for \emph{Retina-100K} and 1.5046 for \emph{Retina-1M} in terms of $L_{Card}$.) However, there are also some samples that exhibit a high label co-occurrence (e.g., it is counted that 5,326 samples exist more than five diseases in \emph{Retina-1M}).

The sorted classes are divided into \textit{many}, \textit{medium}, and \textit{few} groups. The cut-off points and the number of samples ratio are \textit{many} /  \textit{medium} / \textit{few} $\approx$ 100 / 50 / 10 for \textit{Retina-100K}; \textit{many} /  \textit{medium} / \textit{few} $\approx$ 1000 / 200 / 10 for \textit{Retina-1M}.

\begin{figure}[t]
	\includegraphics[width=8.5cm]{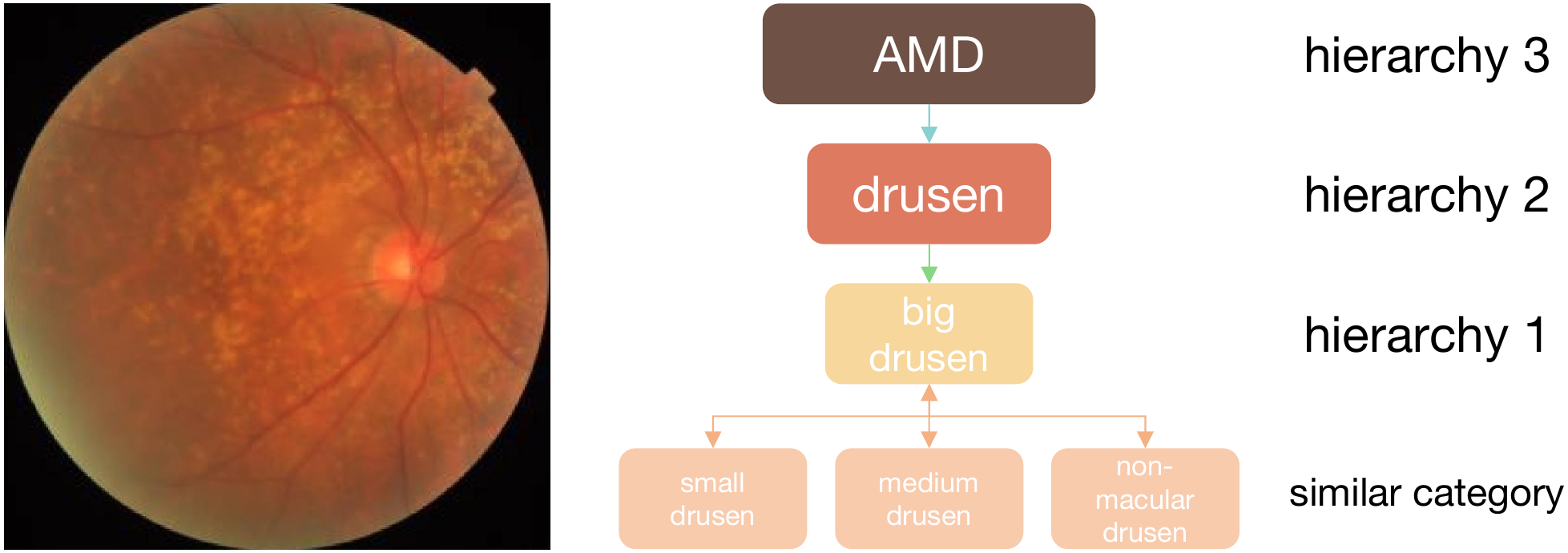}
	\centering
\caption{Each disease is mapped into three hierarchies. e.g., AMD $\rightarrow$ drusen $\rightarrow$ big drusen, with three similar categories from the same higher hierarchy - drusen. We leverage this pre-defined hierarchical information as prior knowledge for training more generalized feature representations.} \label{fig_hierarchy}
\end{figure}

Like the two public datasets mentioned above, all diseases are mapped into a semantic hierarchical tree. Specifically, we regard the 50+ kinds of diseases as the base/finest category, and we divide them into several groups according to their characteristics. There are three hierarchical levels for each base category in total, as shown in Fig.~\ref{fig_hierarchy}. For instance, a sample is labeled as \emph{big drusen}, and it also belongs to \emph{drusen} and \emph{AMD}. There are also three other base categories in the \emph{drusen} subset: \emph{small drusen}, \emph{medium drusen}, and \emph{non-macular drusen}. 
For a better overview of how our hierarchies are conducted and further reproducibility, we give a detailed description of the group generation for different hierarchies in our supplementary files.

\section{Methodology}

\begin{figure*}[t]
	\includegraphics[width=18cm]{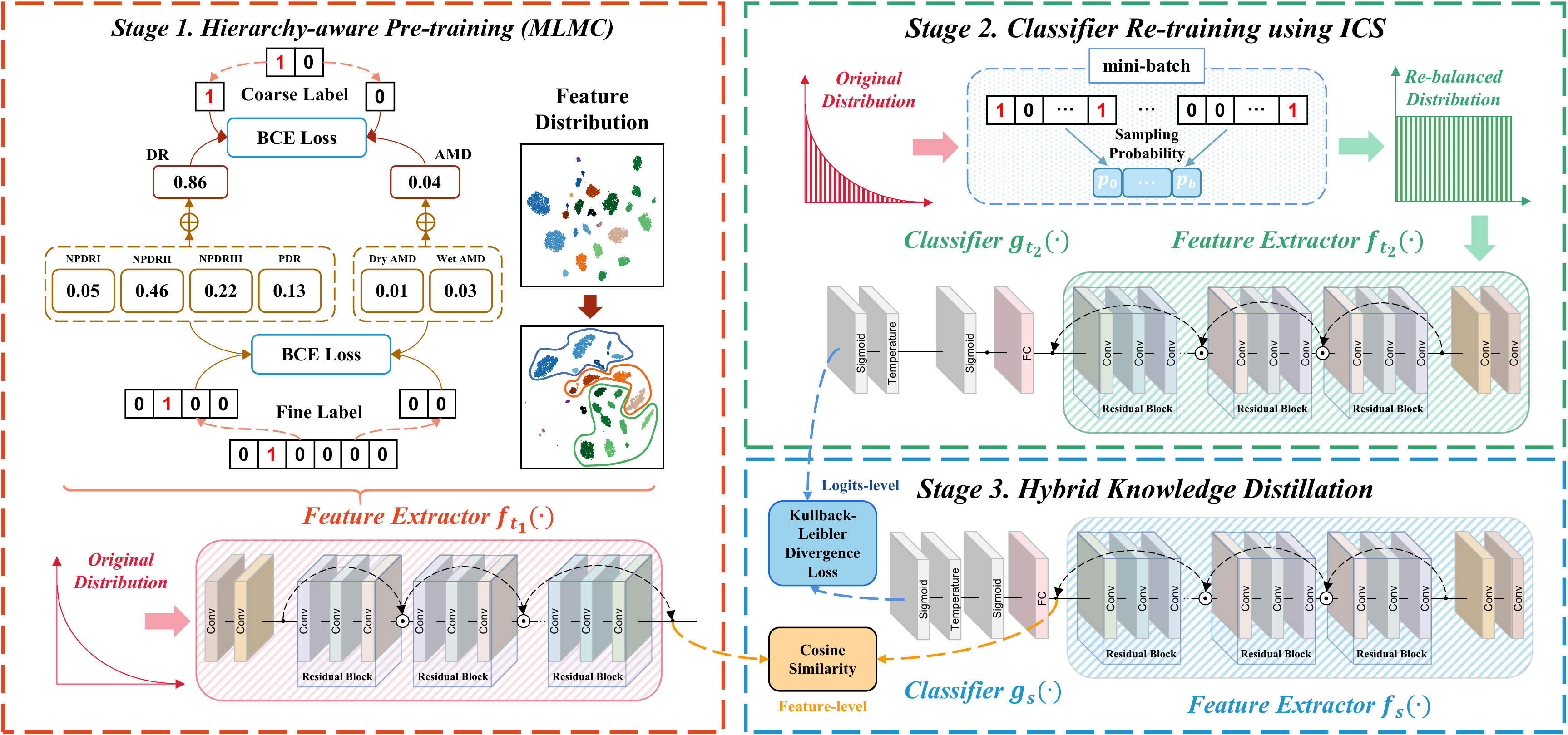}
	\centering
\caption{The overall framework of our proposed methods. The framework consists of three key components. First, we use the hierarchical information of retinal diseases as the prior knowledge for the pre-training of the model $g_{t_{1}}(f_{t_{1}}(\cdot))$. Then, an instance-wise class-balanced loss is introduced to help train the model $g_{t_{2}}(f_{t_{2}}(\cdot))$ which tends to pay more attention to those rare diseases in re-sampling distribution. Finally, the model $g_{t_{1}}(f_{t_{1}}(\cdot))$ with feature-balanced representation and $g_{t_{2}}(f_{t_{2}}(\cdot))$ with fairer classifier are both distilled into a unified student model under a hybrid knowledge distillation manner. } \label{fig_framework}
\end{figure*}
In this section, we first define basic notations for long-tailed multi-label classification in retinal disease recognition. In Sec.~\ref{Sec. hierarchy}, we introduce a novel hierarchy-aware pre-training, which leverages the pre-defined hierarchy of retinal diseases for more efficient representations training. In Sec.~\ref{Sec. ICS}, we analyze how the naive resampling strategies can fail in a multi-label setting. Then we present an instance-wise class-balanced sampling technique to address this challenge. Finally, we propose to use hybrid multiple knowledge distillation to bridge the gap for feature representation and classifier bias following the two-stage methods~\cite{zhou2020bbn,kang2019decoupling}. The overall training process is outlined in Algorithm~\ref{algorithm_1}.

\begin{algorithm}[t]
\caption{HKGL for Retinal Diseases Recognition}\label{alg:cap}

\KwIn{Original data distribution $\mathcal{S} = {(\mathcal{X}, \mathcal{Y})}$, training epochs $E$, hierarchy levels $M$, batch size $b$, teacher model 1 parameters $w_{t_{1}}$, teacher model 2 parameters $w_{t_{2}}$, student model parameters $w_{s}$ learning rate $\eta$, trade-off parameters $\alpha, \beta, \lambda$.}
\emph{\textbf{1. Hierarchy-aware pre-training}}\;
\For{$e = 1, 2, ..., E$}{ 
$\mathcal{{L}}_{all} = 0$\;
\For{$m = 1, 2, ..., M-1$}{
    \textbf{Compute} the logits of corresponding parent categories in level $m+1$ using Equation (2)\;
    \textbf{Compute} the loss of corresponding parent categories ${c}_{k}$ in level $m+1$ using Equation (3)\;
    \textbf{Obtain} loss for the parent classes: $\mathcal{L}_{all}$ += $\mathcal{L}_{BCE}^{m+1}$\;
}
    \textbf{Compute} loss for the classes of the first level using Equation (4): $\mathcal{L}_{all}$ += $\mathcal{L}_{BCE}^{1}$\;
    \textbf{Update} $w_{t_{1}}$ = $w_{t_{1}}$ - $\eta\bigtriangledown \mathcal{{L}}_{all}$\;
}
\emph{\textbf{2. Classifier Re-training using ICS}}\;
\textbf{Freeze} feature extractor (\textit{Optional})\;
$\mathcal{{L}}_{all} = 0$\;
\For{$e = 1, 2, ..., E$}{ 

\textbf{Compute} the loss from the re-balanced distribution using Equation (6)\;
\textbf{Compute} $\mathcal{{L}}_{all}$ += $\mathcal{L}_{BCE}$\;
}
\textbf{Update} $w_{t_{2}}$ = $w_{t_{2}}$ - $\eta\bigtriangledown \mathcal{{L}}_{all}$\;
\emph{\textbf{3. Hybrid Knowledge Distillation}}\;
\For{$e = 1, 2, ..., E$}{ 
    \textbf{Compute} feature-level distillation loss $\mathcal{L}_{F-KD}$ between $w_{t_{1}}$ and $w_{t_{2}}$ using Equation (7)\;
    \textbf{Compute} logits-level distillation loss $\mathcal{L}_{C-KD}$ between $w_{t_{1}}$ and $w_{t_{2}}$ using Equation (8)\;
    \textbf{Compute} classification loss $\mathcal{L}_{BCE}$ using Equation (6)\;
    \textbf{Compute} the overall loss $\mathcal{L}_{total} = (1-\alpha-\beta)\mathcal{L}_{BCE\cdot} + \alpha (\gamma \mathcal{L}_{F-KD}) +\beta \mathcal{L}_{C-KD}$\;
}
\textbf{Update} $w_{s}$ = $w_{s}$ - $\eta\bigtriangledown \mathcal{{L}}_{total}$\;
\KwOut{Student model parameters $w_{s}$.}
\label{algorithm_1}
\end{algorithm}

\subsection{Problem Definition}
 Suppose the original data distribution $\mathcal{S} = \{(\mathcal{X}, \mathcal{Y})\}$, where $\mathcal{X} = \{x_{1}, x_{2}, ..., x_{N}\}$ are the N instances, and $\mathcal{Y} = \{y_{1}, y_{2}, ..., y_{N}\}$ denotes the associated labels. Specially, for a multi-label setting, each sample $\textbf{y}_{i} = [y_{c_{1}}, y_{c_{2}}, ..., y_{c_{k}}]$, where $k$ denotes the total number of possible categories and $\textbf{y}_{i} \in \{0,1\}^{k}$, i.e., $y_{c_{j}}$ = 1 indicates the presence of the category $c_{j}$ in image $x_{i}$ and $y_{c_{j}}$ = 0 otherwise. All indexes of category are sorted into $\{c_{1}, c_{2}, ..., c_{k}\}$ in decreasing order with $N_{c_{1}}>>N_{c_{k}}$. Our goal is to improve the training of a deep neural network model (feature extractor $f(\cdot)$ and classifier $g(\cdot)$) from the long-tailed fundus database for the recognition of various retinal diseases.

\subsection{Pre-Training with Hierarchical Information}
\label{Sec. hierarchy}
\begin{figure}[t]
	\includegraphics[width=8.5cm]{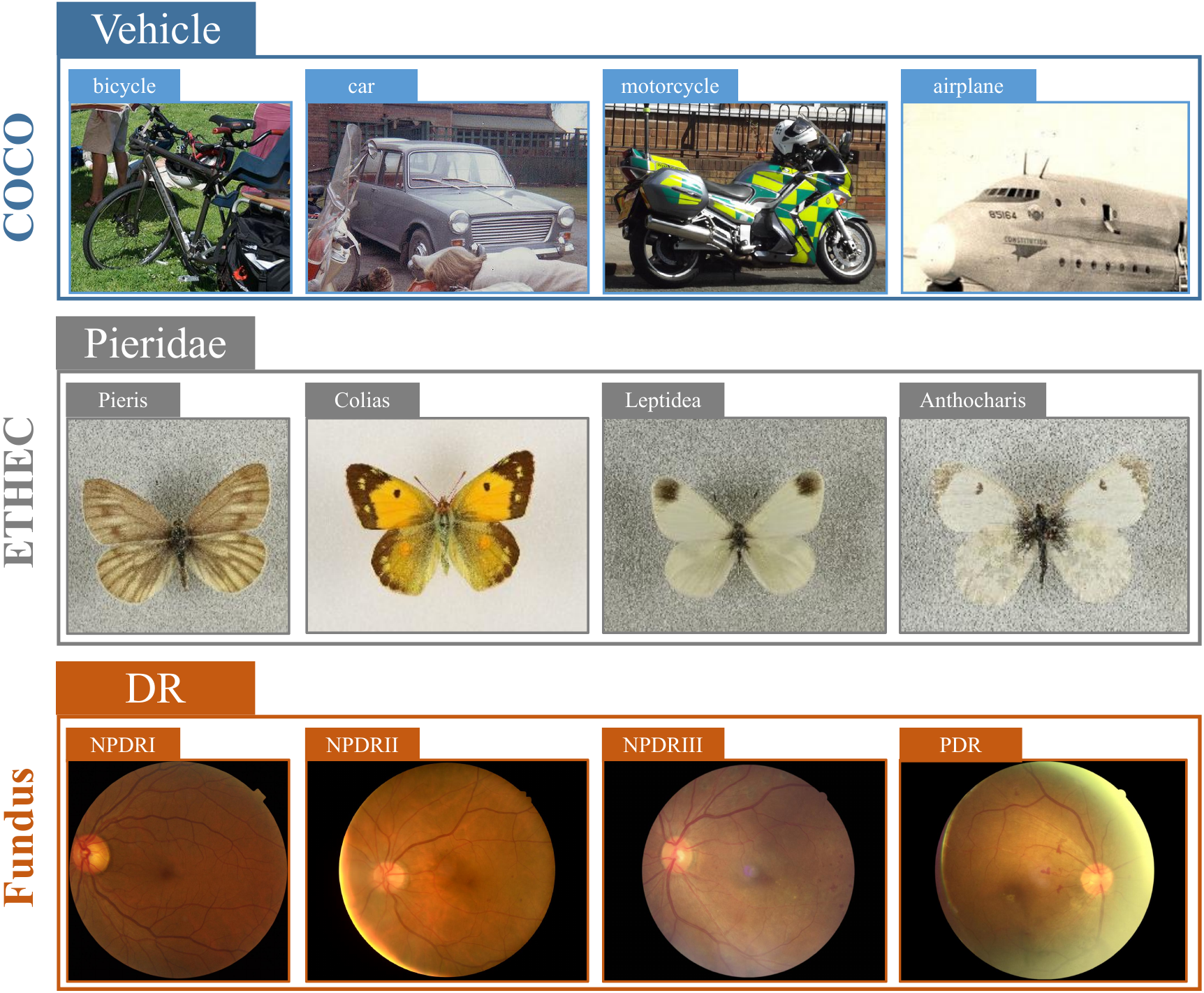}
	\centering
\caption{Three datasets COCO~\cite{lin2014microsoft}, ETHEC~\cite{dhall2020hierarchical} and Kaggle DR+~\cite{ju2022improving}, which define the class-hierarchy. Compared with the two others, sub-classes in some retinal diseases such as DR show a higher semantic similarity.} \label{fig_3datasets}
\end{figure}
The classes of objects in the real world naturally exhibit a hierarchical form, e.g., from domains to species in taxonomy. For instance, we show three datasets both of which define a hierarchy-tree structure for various categories. Here, we define the categories with the finest semantic meaning as \textbf{child class}, some child classes with common features or characteristics in semantics can be grouped into a coarse class, defined as \textbf{parent class}. For example, in COCO~\cite{lin2014microsoft}, the parent class \textbf{Vehicle} consists of several child classes such as \textit{bicycle} and \textit{car}. However, those classes share only a little semantic similarity. ETHEC~\cite{dhall2020hierarchical} is a widely-used dataset for hierarchical classification. Although \textit{Pieris} and \textit{colia} belong to the family \textbf{Pieridae}, it is still difficult to connect them visually. 

Compared with natural images, most retinal diseases can be divided into several subclasses as the lesions progress, such as diabetic retinopathy. Therefore, using the hierarchical information to train the retinal disease diagnosis model shows significant advantages. Different from previous work~\cite{ju2021relational}, which presented 3 kinds of 2-level relational subsets generation: \emph{shot-based}, \emph{region-based} and \emph{feature-based}. In this study, after the discussion with more than 10 senior ophthalmologists, we refined the hierarchical mapping by taking both region information and feature information into consideration, as we presented in Sec. III-Datasets\footnote{We have included a detailed description of hierarchical mapping in our supplementary files.}. In the following parts, we will give a detailed explanation of how to 
incorporate the defined hierarchical information into the training of deep neural networks. 

\begin{figure}[t]
	\includegraphics[width=8.5cm]{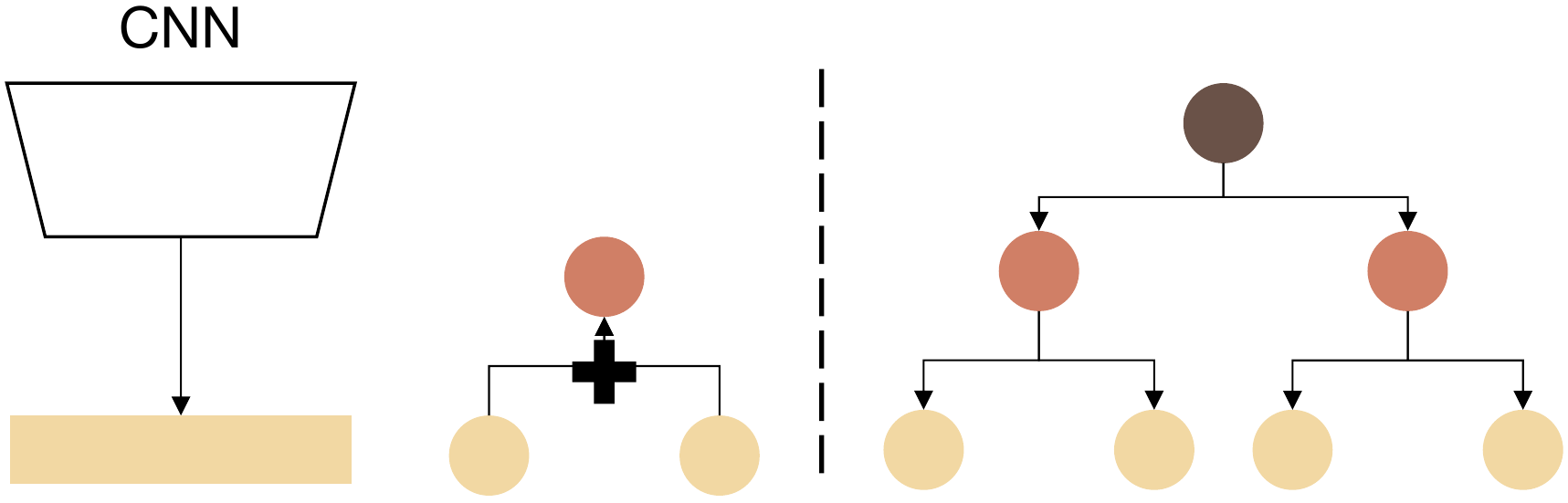}
	\centering
\caption{The illustration of Multi- label Marginalization Classifier (MLMC). We do not directly predict the probability distribution for each level. The probability for parent nodes is determined by summing the logits of their corresponding child nodes.} \label{fig_mlmc}
\end{figure}

Our main idea is to minimize the classification loss of all categories over different hierarchical levels. Assuming we have $M$ hierarchical levels, our target loss could be:
\begin{equation}
    \mathcal{L}_{target} = \sum_{m=1}^{M}\mathcal{L}_{BCE}^{m},
\end{equation}
where BCE denotes the binary cross-entropy loss. A potential solution for hierarchy-aware constraints could be per-level classifiers~\cite{dhall2020hierarchical}. The model explicitly outputs separate predictions per level for a given image using $\mathcal{M}$ classifiers. However, per-level classifiers lose connection across different hierarchical levels since the parameters of each classifier are optimized independently.

In this study, we introduce Multi-label Marginalization classifier (MLMC) as an extension of the marginalization classifier~\cite{dhall2020hierarchical}, which is shown in Fig.~\ref{fig_mlmc},  to inject the hierarchical information of retinal diseases into the model training. Formally, a single classifier $g(\cdot)$ outputs a probability distribution over the final level (level 1) of the class hierarchy (finest classes). Instead of building classifiers for the remaining parent classes of $M - 1$ levels, we compute the probability distribution over each one by summing the probability of their corresponding child classes. 

Given the parent category ${c}_{k}$ in level $m+1$ and its corresponding $o_{m}$ child classes $\{c_{1}, c_{2}, ..., c_{o_{m}}\}$ in level $m$, we can obtain its predicted probabilities for level $m$:
\begin{equation}
    \hat{\mathcal{Y}}_{c_{k}}^{m+1} = \sigma (\sum_{j=1}^{o_{m}}\hat{\mathcal{Y}}_{c_{j}}^{m}),
\end{equation}
where $\sigma(\cdot)$ denotes the sigmoid activation function. Then, we calculate the loss of parent category ${c}_{k}$ using general BCE loss:
\begin{equation}
    \mathcal{L}_{BCE}^{m+1} = \mathcal{L}_{BCE}(\hat{\mathcal{Y}}_{c_{k}}^{m+1}, \mathcal{Y}^{m+1}).
\end{equation}
It should be noted that, for the prediction probabilities of the lowest  hierarchy $l_{1}$, we directly use the logits from the outputs of fully-connected layer $g(f(\mathcal{X}))$:
\begin{equation}
    \mathcal{L}_{BCE}^{1} = \mathcal{L}_{BCE}(\sigma(g(f(\mathcal{X}))), \mathcal{Y}^{1}).
\end{equation}
Then, losses across all levels are summed for global optimization. Compared with per-level classifiers, although MLMC does not explicitly predict scores of parent classes, the models is still penalized for incorrect predictions across the $M$ levels. Also, as we mentioned in Sec.~\ref{sec. rfmid}, for those child classes that could belong to more than one parent class, MLMC still has good generalization ability by simply summing the logits of the corresponding classes to different parent classes.

\subsection{Instance-wise Class-balanced Sampling}
\label{Sec. ICS}
\begin{figure}[t]
	\includegraphics[width=9cm]{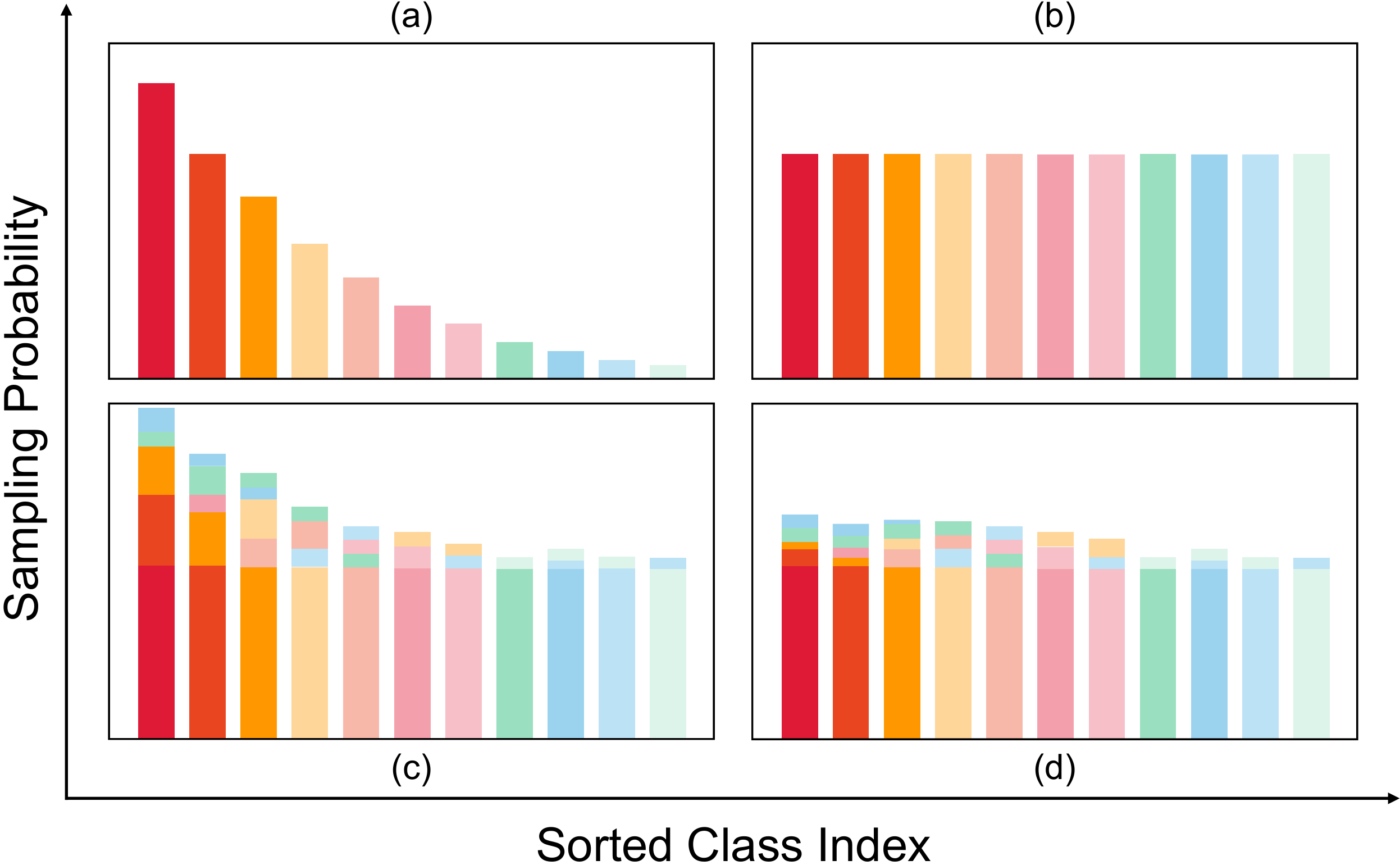}
	\centering
\caption{The illustration of sampling probability under different sampling strategies: (a) instance-balanced sampling; (b) expected sampling; (c) class-balanced sampling; (d) instance-wise class-balanced sampling.} \label{fig_sampling}
\end{figure}
\label{sec. sampling}
In this section, we introduce two commonly used sampling strategies in the long-tailed multi-class classification: \emph{instance-balanced sampling} and \emph{class-balanced sampling}. Then we give an experimental analysis to explain why these two sampling ways do not work well in a multi-label setting. Finally, we propose to use an instance-wise class-balanced sampling to handle this scenario.
\subsubsection{Instance-balanced Sampling}
In the multi-label classification, we always train a deep neural network by minimizing a binary cross-entropy (BCE) loss:
\begin{equation}
    \mathcal{L}_{BCE} = \frac{1}{kN}\sum_{i=1}^{N}\sum_{j=1}^{k}[y_{i}^{j}log(\hat{y}_{i}^{j}) + (1-y_{i}^{j})log(1-\hat{y}_{i}^{j})],
\end{equation}
where $\hat{y}_{i}^{j}$ denotes the predicted results for the $i$-th sample in the class of $c_{j}$. It can be seen that each example $x_{i} \in \mathcal{X}$ obtains the same sampling probability in a mini-batch during the training. However, the head classes will be sampled more frequently than the
tail classes in a long-tailed distribution dataset and the model tends to under-fit those classes with fewer samples, which results in a prediction bias.
\subsubsection{Class-balanced Sampling}
Given the sampling probability $p_{i}$ for each sample in a mini-batch, and we have the $p_{i} = \frac{N_{c_{j}}}{N}$ in an instance-balanced sampling. Instead, class-balanced sampling aims to assign an equal sampling probability $p_{c_{k}} = \frac{1}{k}$ for each category in a mini-batch (See Fig.~\ref{fig_sampling}-(b)). Class-balanced sampling is a simple but effective trick for training models on imbalanced data and has been demonstrated to be a necessary component in many state-of-the-art works~\cite{zhou2020bbn,kang2019decoupling}. However, there are two disadvantages, especially for a multi-label setting. First, under an ideal condition, we have $K$ examples from $K$ kinds of categories in a mini-batch with a batch size of $K$. In this case, the instance-level sampling probability becomes $p_{i}^{j} = \frac{p_{c_{j}}}{N_{j}}$. Since $N_{c_{1}}>>N_{c_{k}}$, those examples from the head classes will be less exposed to the model during the training phase, so the learned feature space is incomplete. Second, due to the label co-occurrence, over-sampling an instance 
that belongs to both of a head class and a tail class can bring a new relative imbalance issue, as shown in Fig.~\ref{fig_sampling}-(c).

\subsubsection{Instance-wise Class-balanced Sampling}
As illustrated in Fig.~\ref{fig_sampling}-(c), wrongly sampling those samples with label co-occurrence can release new relative imbalance. To address the above limitations of sampling multi-label data, an intuitive solution is to 'ignore' those samples covering too many categories and avoid sampling such samples as much as possible. Hence, we present the instance-wise class-balanced sampling (ICS) strategy as~\cite{wu2020distribution,guo2021learning}. Since we have the expected ideal sampling probability for one sample from category $c_{j}$: $p_{i}^{j} = \frac{p_{c_{j}}}{N_{c_{j}}}$ and the actual sampling probability: $p_{i}^{A} = \sum_{j=1}^{k}\mathbbm{1}_{j}\cdot p_{i}^{j}$. Then, a sampling factor $\delta$ is used to re-balanced the sampling probability $\delta_{i}^{j} = \frac{p_{i}^{j}}{p_{i}^{A}}$, then the BCE loss becomes:
\begin{equation}
    \mathcal{L}_{BCE} = \frac{1}{kN}\sum_{i=1}^{N}\sum_{j=1}^{k}[y_{i}^{j}log(\hat{y}_{i}^{j}) + (1-y_{i}^{j})log(1-\hat{y}_{i}^{j})] \cdot \sqrt{\delta_{i}^{j}}.
\end{equation}

However, we find that some mild diseases which are also in the head class, such as tessellated fundus, exist in more than half of the samples from other categories, which push the sampling factor $\delta_{i}^{j}$ close to zero and complicates the optimization. Unlike~\cite{wu2020distribution} using two hyper-parameters to map the sampling factor near to one, we directly square it so as to increase $\delta_{i}^{j}$ rapidly, and have the sampling probability for those samples with different co-occurrence degrees discriminative. 
Hereafter, we indicate ICS as $L_{BCE}$ for simplification.

\subsection{Hybrid Multiple Knowledge Distillation}
Xiang et al.~\cite{xiang2020learning} and Ju et al.~\cite{ju2021relational} have explored knowledge distillation in long-tailed classification. The original long-tailed distribution is divided into several subsets, which are in a relatively balanced status. Those subsets are used to train several teacher models, which are then distilled into a unified student model. However, there are some obvious shortcomings: (1) the divided subsets result in an incomplete distribution and limited features representation learned by the teacher model which may constraint the performance of the student model; (2) most performance improvements benefit from the knowledge distillation instead of learning from a relatively balanced subset; (3) some rare diseases can not directly be divided by independent feature-based or region-based rules proposed by~\cite{ju2021relational}. 

Another observation is that, two-stage learning methods~\cite{kang2019decoupling, zhang2021distribution} indicates that \textit{``training from imbalanced distribution produces a strong feature representation and the instance-based sampling produces a less-biased classifier."} To this end, we propose to enhance the two-stage learning and multiple knowledge distillation by introducing a novel hybrid multiple knowledge distillation method which can distill efficient information for long-tailed learning on both feature-level and classifier-level, i.e., logits-level~\cite{Hinton2015Distilling}.

Formally, we first trained two teacher models $\mathcal{F}_{t_{1}} = g_{t_{1}}(f_{t_{1}}(\cdot))$ and $\mathcal{F}_{t_{2}} = g_{t_{2}}(f_{t_{2}}(\cdot))$ using general classification loss (e.g., MLMC) and re-sampling classification loss (e.g., ICS-MLMC). Inspired by two-stage training~\cite{zhou2020bbn,kang2019decoupling, guo2021long} that model learns a good representation for feature extractor under an original distribution, we first leverage the feature distillation from $\mathcal{F}_{t_{1}}$ to assist the student model training. Given the student model $\mathcal{F}_{s} = g_{s}(f_{s}(\cdot))$, The feature-level knowledge distillation can be formulated as follows:

\begin{equation}
    \mathcal{L}_{F-KD} = l_{sim}(v_{t_{1}},v_{s}),
\label{eq. fkd}
\end{equation}
where the extracted feature $v = f(x)$, and $l_{sim}(\cdot) = 1 - cos(\cdot)$ to calculate the cosine distance (similarity) between output features from teacher and student models. Then, the standard KD is used to distill the knowledge from $F_{t_{2}}$ to the student model, which can be formulated as:

\begin{equation}
    \mathcal{L}_{C-KD} = l_{KD}(\sigma(g_{s}(f_{s}(\mathcal{X})))/\mathcal{T}, \sigma(g_{t_{2}}(f_{t_{2}}(\mathcal{X}))/\mathcal{T})),
\label{eq. tkd}
\end{equation}
where $\mathcal{T}$ is the hyper-parameter for temperature scaling and $l_{KD}$ is the Kullback-Leibler divergence loss: $l_{KD}(p_{1}|p_{2}) = p_{1}\cdot \mathrm{ln}\frac{p_{1}}{p_{2}}$. Hence, we have the total loss:

\begin{equation}
    \mathcal{L}_{total} = (1-\alpha-\beta)\mathcal{L}_{BCE\cdot} + \alpha (\gamma \mathcal{L}_{F-KD}) +\beta \mathcal{L}_{C-KD},
\label{eq. total_loss}
\end{equation}
where $\alpha$ and $\beta$ are used to control the KD loss weights while $\gamma$ for prevent $L_{F-KD}$ from being close to zero. For the simplification of the hyper-parameters searches, we keep $\alpha = \beta$ here.

Hybrid multiple knowledge distillation enables the teacher model to train from the original distribution using a traditional knowledge distillation manner~\cite{Hinton2015Distilling}. In this way, teacher models maintain the complete feature representations, to prevent the student model's performance from being limited by itself performance bottlenecks.

\section{Experiments}
\subsection{Implementation Details}
We use ResNet-50~\cite{he2016deep} with pre-trained weights from ImageNet as our backbone network. The input size is 512 × 512 for two in-house datasets and 224 × 224 for two public datasets, respectively. We apply Adam to optimize the model. The learning
rate starts at $1 \times 10^{-3}$ and decreases ten-fold when there is no drop in validation
loss till $1 \times 10^{-7}$ with the patience of 5 epochs. The $\alpha$ and $\beta$ are set as 0.2. The $\gamma$ is set as 10. We apply regular data-augmentation transformations during the training phase, such as random crop and flip. Following most long-tailed multi-label works~\cite{wu2020distribution,guo2021long}, we use the mean average precision (mAP) as the evaluation metric. All the results are calculated after 5-time running with random seeds and a batch size of 128. In all results presented, we report the 5-trial average performance and mean standard deviation. All experiments are implemented using the PyTorch platform and 8 × NVIDIA RTX 3090 GPUs.

\begin{table*}[t]
\centering

\normalsize
\caption{The performance (mAP) of the proposed method and comparison methods on two datasets \textit{Retina-100K} and \textit{Retina-1M}.} \begin{threeparttable}
\resizebox{18cm}{2.8cm}{\begin{tabular}{c|cccc|cccc}
\hline
\hline
Dataset                                     & \multicolumn{4}{c|}{Retina-100K}                                                   & \multicolumn{4}{c}{Retina-1M}                                                      \\ \hline \hline
Methods                                     & many  & medium & \multicolumn{1}{c|}{few}   & \multicolumn{1}{c|}{average} & many  & medium & \multicolumn{1}{c|}{few}   & \multicolumn{1}{c}{average} \\ \hline 
ERM                                         & 70.89 ($\pm$0.13) & 71.93 ($\pm$1.26) & \multicolumn{1}{c|}{35.90 ($\pm$3.74)} & \multicolumn{1}{c|}{59.57 ($\pm$1.71)}     & 84.70 ($\pm$2.06) & 61.57 ($\pm$1.85)  & \multicolumn{1}{c|}{41.83 ($\pm$2.96)} & \multicolumn{1}{c} {62.70 ($\pm$2.29)}    \\  \hline
RS                                          & 65.72 ($\pm$2.12) & 67.17 ($\pm$0.93)  & \multicolumn{1}{c|}{36.99 ($\pm$1.21)} & \multicolumn{1}{c|}{56.63  ($\pm$1.85)}     & 80.66 ($\pm$0.82) & 58.45 ($\pm$0.51)  & \multicolumn{1}{c|}{42.72 ($\pm$1.73)} & \multicolumn{1}{c}{60.61 ($\pm$1.02)}    \\
RW                                          & 71.32 ($\pm$0.70) & 73.11 ($\pm$4.67)  & \multicolumn{1}{c|}{39.03 ($\pm$3.68)} & \multicolumn{1}{c|}{61.15 ($\pm$3.02)}     & 84.02 ($\pm$1.45) & \textcolor{blue}{\textbf{63.98}} ($\pm$3.31)  & \multicolumn{1}{c|}{43.69 ($\pm$4.40)} & \multicolumn{1}{c}{63.90 ($\pm$3.05)}    \\  \hline

OLTR~\cite{liu2019large}                    & 70.22 ($\pm$1.52) & 72.08 ($\pm$0.60)  & \multicolumn{1}{c|}{39.00 ($\pm$0.88)} & \multicolumn{1}{c|}{60.43 ($\pm$1.00)}     & 80.10  ($\pm$0.12) & 62.31 ($\pm$0.35)  & \multicolumn{1}{c|}{42.01 ($\pm$0.49)} & \multicolumn{1}{c}{61.47 ($\pm$0.32)}    \\

RSKD~\cite{ju2021relational}                & 70.65 ($\pm$2.27) & \textcolor{blue}{\textbf{73.98}} ($\pm$1.38)  & \multicolumn{1}{c|}{\textcolor{blue}{\textbf{41.56}} ($\pm$0.47)}  & \multicolumn{1}{c|}{\textcolor{blue}{\textbf{62.06}} ($\pm$1.38)}     & 80.22 ($\pm$0.57) & 63.88 ($\pm$0.40)  & \multicolumn{1}{c|}{\textcolor{blue}{\textbf{44.10}} ($\pm$2.39)} & \multicolumn{1}{c}{62.73 ($\pm$1.12)}    \\
\hline

Focal Loss~\cite{lin2017focal}              & 72.84 ($\pm$1.97) & 73.37 ($\pm$1.42)  & \multicolumn{1}{c|}{39.13 ($\pm$2.23)} & \multicolumn{1}{c|}{61.78 ($\pm$1.87)}    & 85.74 ($\pm$1.52) & 62.15 ($\pm$0.05)  & \multicolumn{1}{c|}{42.99 ($\pm$0.59)} & \multicolumn{1}{c}{63.63 ($\pm$0.72)}    \\

LDAM~\cite{cao2019learning}                 & 71.24 ($\pm$3.54) & 73.75 ($\pm$3.68)  & \multicolumn{1}{c|}{39.12 ($\pm$0.86)} & \multicolumn{1}{c|}{61.37 ($\pm$2.70)}     & 85.62 ($\pm$0.10) & 63.28 ($\pm$0.69)  & \multicolumn{1}{c|}{42.05 ($\pm$1.14)} & \multicolumn{1}{c}{63.65 ($\pm$0.64)}    \\

CBLoss-Focal~\cite{cui2019class}            & 51.93 ($\pm$1.30) & 50.66 ($\pm$3.63)  & \multicolumn{1}{c|}{20.79 ($\pm$0.42)} & \multicolumn{1}{c|}{41.13 ($\pm$1.78)}     & 77.14 ($\pm$0.78) & 50.69 ($\pm$2.69)  & \multicolumn{1}{c|}{20.66 ($\pm$1.10)} & \multicolumn{1}{c}{49.50 ($\pm$1.53)}    \\

DBLoss-Focal~\cite{wu2020distribution}      & 72.61 ($\pm$1.63) & 72.39 ($\pm$0.53)  & \multicolumn{1}{c|}{38.59 ($\pm$0.08)} & \multicolumn{1}{c|}{61.27 ($\pm$0.74)}     & \textcolor{red}{\textbf{85.99}} ($\pm$3.08) & 62.15 ($\pm$2.08)  & \multicolumn{1}{c|}{43.22 ($\pm$1.26)} & \multicolumn{1}{c}{63.79 ($\pm$2.14)}    \\

ASL~\cite{ben2020asymmetric}                & \textcolor{blue}{\textbf{72.94}} ($\pm$0.56) & 73.67 ($\pm$1.40)  & \multicolumn{1}{c|}{39.21 ($\pm$1.63} & \multicolumn{1}{c|}{61.94 ($\pm$1.20)}      & 85.10 ($\pm$2.60) & 63.79 ($\pm$2.43)  & \multicolumn{1}{c|}{43.45 ($\pm$1.29)}  & \multicolumn{1}{c}{\textcolor{blue}{\textbf{64.11}} ($\pm$2.11)}    \\ \hline

baseline-original                           & 71.18 ($\pm$1.32) & 72.33 ($\pm$2.01)  & \multicolumn{1}{c|}{37.39 ($\pm$2.35)} & \multicolumn{1}{c|}{60.29 ($\pm$1.89)}     & 85.02 ($\pm$2.56) & 62.27 ($\pm$1.03)  & \multicolumn{1}{c|}{42.17 ($\pm$1.91)} & \multicolumn{1}{c}{63.15 ($\pm$1.83)}    \\

baseline-ICS                                & 70.67 ($\pm$1.03) & 73.40 ($\pm$1.91)  & \multicolumn{1}{c|}{40.95 ($\pm$2.08)} & \multicolumn{1}{c|}{61.67 ($\pm$1.67)}     & 84.86 ($\pm$1.12) & 62.39 ($\pm$2.56)  & \multicolumn{1}{c|}{43.44 ($\pm$1.78)} & \multicolumn{1}{c}{63.56 ($\pm$1.82)}    \\ \hline

Ours                                        & \textcolor{red}{\textbf{73.86}} ($\pm$1.23) & \textcolor{red}{\textbf{74.75}} ($\pm$0.96)  & \multicolumn{1}{c|}{\textcolor{red}{\textbf{43.82}} ($\pm$1.36)} & \multicolumn{1}{c|}{\textcolor{red}{\textbf{64.14}} ($\pm$1.19)}    & \textcolor{blue}{\textbf{85.79}} ($\pm$1.64) & \textcolor{red}{\textbf{64.00}} ($\pm$0.92)  & \multicolumn{1}{c|}{\textcolor{red}{\textbf{44.28}} ($\pm$1.38)} & \multicolumn{1}{c}{\textcolor{red}{\textbf{64.69}} ($\pm$1.32)}    \\ \hline
\hline
\end{tabular}}
\begin{tablenotes}
\item[1] The sorted classes are divided into \textit{many}, \textit{medium} and \textit{few} groups and mAPs are calculated accordingly. 
\item[2] For the global evaluation, we report the average performance of three groups (denoted by \textbf{average}).
\item[3] The best and second best performance are marked in \textcolor{red}{\textbf{red}} and \textcolor{blue}{\textbf{blue}}, respectively.
\end{tablenotes}

\end{threeparttable}
\label{Table_sota}

\end{table*}

\subsection{Comparison Study on Two in-house Datasets}
In this section, we give a comprehensive comparison study on two in-house datasets with various baselines including some state-of-the-art works for long-tailed classification: (1) \textbf{Empirical Risk Minimization (ERM)};  (2) vanilla \textbf{Re-sampling (RS)} ; (3) vanilla \textbf{Re-weighting (RW)}; (4) \textbf{OLTR}~\cite{liu2019large}; (6) \textbf{Focal Loss}~\cite{lin2017focal} ; (7) \textbf{LDAM}~\cite{cao2019learning}; (8) \textbf{CBLoss}~\cite{cui2019class} (9) \textbf{DBLoss}~\cite{wu2020distribution}; (10) \textbf{ASL}~\cite{ben2020asymmetric}.

The overall results are shown in Table~\ref{Table_sota}.
Different baseline methods are grouped by their model designs, such as loss functions. The results are presented by groups \textit{many, medium } and \textit{few} to test how each method reacts to classes with different numbers of samples. In the following, we also use the head and tail classes to refer to group many and group medium. We report the average results of the three groups to view the global performance, which is denoted as ``average". To better understand the trade-off between different shot-based groups for different comparison methods, we mark the best and second performance in red and blue, respectively. From the results reported in Table~\ref{Table_sota}, among all baselines, ASL achieves the best results - 61.94\% mAP of average results over three groups and 52.15\% mAP over all classes on \textit{Retina-100K}, followed by Focal Loss - 61.78\% mAP. 
Moreover, we find that although RS achieves a good performance in the group of few, it brings a catastrophic impact on the many-shot classes (70.89\% mAP $\rightarrow$ 65.72\% mAP) and medium-shot classes (71.93\% mAP $\rightarrow$ 67.17\% mAP), resulting from the label co-occurrence in a multi-label setting. 
However, we fail to obtain a satisfactory performance in terms of CBLoss, which was originally designed for multi-class long-tailed classification. 

For our framework, we first train two vanilla teacher models using different sampling strategies under hierarchy-aware constraints as teacher models, denoted by ``baseline-original" and ``baseline-ICS", respectively. 
Then we distill the knowledge from two baseline teacher models into a unified student model under a hybrid distillation manner. Take~\textit{Retina-100K} for instance, the results of ``baseline-original" show that the hierarchical pre-training can benefit the ERM baseline model (from 59.57 \%mAP to 60.29 \%mAP) without using any specific re-sampling strategies. 
Furthermore, the ICS can further improve the overall performance (from 60.29 \%mAP to 61.67 \%mAP) with a marginal performance loss in the many-shot group setting (from 71.18 \%mAP to 70.67 \%mAP). The last row demonstrates the superior performance of our proposed methods, which outperform all competitors on both two datasets \textit{Retina-100K} and \textit{Retina-1M} and the superiority holds for all metrics. 

\begin{table}[]
\centering
\normalsize
\caption{The results of comparison study on RFMiD dataset.}
\resizebox{9cm}{2.4cm}{\begin{tabular}{c|cccc}
\hline \hline
Dataset      & \multicolumn{4}{c}{RFMiD}                             \\ \hline
Methods      & many  & medium & \multicolumn{1}{c|}{few}   & average \\ \hline
ERM          & \textcolor{blue}{\textbf{70.93}} ($\pm$1.31) & 57.89 ($\pm$1.22)  & \multicolumn{1}{c|}{14.85 ($\pm$1.95)} & 47.89 ($\pm$1.49)   \\ \hline
RS           & 68.67 ($\pm$1.23) & \textcolor{red}{\textbf{61.48}} ($\pm$1.73)  & \multicolumn{1}{c|}{\textcolor{blue}{\textbf{25.94}} ($\pm$2.51)} & \textcolor{blue}{\textbf{52.03}} ($\pm$1.83)   \\
RW           & 70.27 ($\pm$1.30) & 60.00 ($\pm$1.79)  & \multicolumn{1}{c|}{18.71 ($\pm$1.60)} & 49.66 ($\pm$1.56)   \\ \hline
OLTR         & \textcolor{red}{\textbf{71.25}} ($\pm$1.13) & 60.22 ($\pm$1.05)  & \multicolumn{1}{c|}{20.77 ($\pm$1.33)} & 50.75 ($\pm$1.17)   \\
RSKD         & 70.55 ($\pm$0.59) & 59.63 ($\pm$0.23)  & \multicolumn{1}{c|}{22.15 ($\pm$1.20)} & 50.78 ($\pm$0.67)   \\ \hline
Focal        & 70.65 ($\pm$1.14) & 55.53 ($\pm$1.61)  & \multicolumn{1}{c|}{16.42 ($\pm$1.92)} & 47.53 ($\pm$1.56)   \\
LDAM         & 46.67 ($\pm$2.28) & 3.19  ($\pm$3.10)   & \multicolumn{1}{c|}{1.18 ($\pm$3.36)}  & 17.01  ($\pm$2.91)   \\
CBLoss-Focal & 67.73 ($\pm$1.30) & 50.89 ($\pm$0.75)  & \multicolumn{1}{c|}{24.65 ($\pm$1.12)} & 47.77 ($\pm$1.06)   \\
DBLoss-Focal & 68.16 ($\pm$2.52) & 55.27 ($\pm$2.78)  & \multicolumn{1}{c|}{18.94 ($\pm$2.49)} & 47.46 ($\pm$2.60)   \\
ASL          & 68.25 ($\pm$1.93) & 58.25 ($\pm$1.76)  & \multicolumn{1}{c|}{19.59 ($\pm$1.67)} & 48.70 ($\pm$1.79)   \\ \hline
Ours         & 70.50 ($\pm$1.25) & 59.95 ($\pm$1.79)  & \multicolumn{1}{c|}{22.83 ($\pm$1.38)} & 51.09 ($\pm$1.47)   \\
Ours (RS)    & 69.75 ($\pm$1.82) & \textcolor{blue}{\textbf{61.26}} ($\pm$1.82)  & \multicolumn{1}{c|}{\textcolor{red}{\textbf{25.98}} ($\pm$1.91)} & \textcolor{red}{\textbf{52.33}} ($\pm$1.85) \\
\hline \hline
\end{tabular}}
\label{table_rfmid}
\end{table}

\begin{table*}[]
\centering
\normalsize
\caption{The results of comparison study on ODIR dataset.}
\resizebox{18cm}{2.8cm}{\begin{tabular}{c|cccccccc}
\hline \hline
Dataset      & \multicolumn{8}{c}{ODIR}                                                                                                 \\ \hline
Split       & \multicolumn{4}{c|}{Off-Site}                                                   & \multicolumn{4}{c}{On-Site}                               \\ \hline
Methods      & many  & medium & \multicolumn{1}{c|}{few}   & \multicolumn{1}{c|}{average} & many  & medium & \multicolumn{1}{c|}{few}   & average \\ \hline
ERM          & 48.47 ($\pm$0.73) & 46.80 ($\pm$0.44)  & \multicolumn{1}{c|}{11.22 ($\pm$0.56)} & \multicolumn{1}{c|}{35.50 ($\pm$0.58)}   & 50.74 ($\pm$1.64) & 36.46 ($\pm$1.86)  & \multicolumn{1}{c|}{12.79 ($\pm$1.49)} & 33.33 ($\pm$1.66)   \\ \hline

RS           & 46.34 ($\pm$1.08) & \textcolor{red}{\textbf{49.27}} ($\pm$1.59)  & \multicolumn{1}{c|}{9.07 ($\pm$1.57)}  & \multicolumn{1}{c|}{34.89 ($\pm$1.41)}   & 47.91 ($\pm$1.92) & \textcolor{red}{\textbf{39.10}} ($\pm$1.53)  & \multicolumn{1}{c|}{15.35 ($\pm$1.95)} & 34.12 ($\pm$1.80)   \\

RW           & \textcolor{red}{\textbf{50.56}} ($\pm$0.32) & 48.12 ($\pm$1.10)  & \multicolumn{1}{c|}{11.57 ($\pm$1.91)} & \multicolumn{1}{c|}{36.75 ($\pm$1.11)}   & 51.39 ($\pm$1.67) & 37.86 ($\pm$2.24)  & \multicolumn{1}{c|}{17.92 ($\pm$1.26)} & 35.72 ($\pm$1.73)   \\ \hline

OLTR         & 47.37 ($\pm$0.87) & 45.02 ($\pm$0.56)  & \multicolumn{1}{c|}{11.86 ($\pm$0.70)} & \multicolumn{1}{c|}{34.75 ($\pm$0.71)}   & 50.11 ($\pm$0.66) & 36.01 ($\pm$1.45)  & \multicolumn{1}{c|}{20.78 ($\pm$2.69)} & 35.63 ($\pm$1.60)   \\

RSKD         & 48.09 ($\pm$0.20) & 47.78 ($\pm$0.82)  & \multicolumn{1}{c|}{10.82 ($\pm$1.44)} & \multicolumn{1}{c|}{35.56 ($\pm$0.82)}   & 48.89 ($\pm$1.20) & \textcolor{blue}{\textbf{38.61}} ($\pm$1.55)  & \multicolumn{1}{c|}{\textcolor{red}{\textbf{31.21}} ($\pm$2.64)} & \textcolor{red}{\textbf{39.57}} ($\pm$1.80)   \\ \hline

Focal        & 46.63 ($\pm$1.76) & 46.89 ($\pm$1.45)  & \multicolumn{1}{c|}{13.32 ($\pm$2.47)} & \multicolumn{1}{c|}{35.61 ($\pm$1.90)}   & 47.92 ($\pm$2.52) & 35.41 ($\pm$2.21)  & \multicolumn{1}{c|}{10.49 ($\pm$2.08)} & 31.27 ($\pm$2.27)   \\

LDAM         & 41.14 ($\pm$2.42) & 8.22 ($\pm$2.94)   & \multicolumn{1}{c|}{0.48 ($\pm$0.11)}   & \multicolumn{1}{c|}{16.61 ($\pm$1.82)}   & 42.97 ($\pm$2.87) & 5.10 ($\pm$2.26)   & \multicolumn{1}{c|}{0.55 ($\pm$0.22)}  & 16.21 ($\pm$1.78)   \\

CBLoss-Focal & 39.30 ($\pm$1.20) & 47.44 ($\pm$2.50)  & \multicolumn{1}{c|}{10.00 ($\pm$1.73)} & \multicolumn{1}{c|}{32.25 ($\pm$1.81)}   & 43.40 ($\pm$1.44) & 32.31 ($\pm$1.29)  & \multicolumn{1}{c|}{8.60 ($\pm$1.47)}  & 28.10 ($\pm$1.40)   \\

DBLoss-Focal & 48.39 ($\pm$1.27) & 47.11 ($\pm$1.41)  & \multicolumn{1}{c|}{\textcolor{blue}{\textbf{27.83}} ($\pm$1.34)} & \multicolumn{1}{c|}{\textcolor{blue}{\textbf{41.11}} ($\pm$1.34)}   & 50.06 ($\pm$0.87) & 37.60 ($\pm$1.07)  & \multicolumn{1}{c|}{12.96 ($\pm$1.27)} & 33.54 ($\pm$1.06)   \\

ASL          & 47.93 ($\pm$1.78) & 47.89 ($\pm$1.50)  & \multicolumn{1}{c|}{18.57 ($\pm$1.13)} & \multicolumn{1}{c|}{38.13 ($\pm$1.47)}   & \textcolor{red}{\textbf{51.69}} ($\pm$0.77) & 37.36 ($\pm$1.46)  & \multicolumn{1}{c|}{23.70 ($\pm$1.79)} & 37.58 ($\pm$1.34)   \\ \hline

Ours         & \textcolor{blue}{\textbf{49.02}} ($\pm$1.29) & \textcolor{blue}{\textbf{48.26}} ($\pm$1.68)  & \multicolumn{1}{c|}{\textcolor{red}{\textbf{28.05}} ($\pm$1.13)} & \multicolumn{1}{c|}{\textcolor{red}{\textbf{41.78}} ($\pm$1.37)}   & \textcolor{blue}{\textbf{51.58}} ($\pm$2.27) & 36.82 ($\pm$1.67)   & \multicolumn{1}{c|}{\textcolor{blue}{\textbf{28.98}} ($\pm$2.21)} & \textcolor{blue}{\textbf{39.12} }($\pm$2.05)   \\ \hline \hline

\end{tabular}}
\label{table_odir}
\end{table*}

\begin{table}[]
\small
\caption{Ablation analysis on different model components.}
\centering
\begin{tabular}{cccc|c|c}
\hline \hline
MLMC                      & ICS                       & cRT            & Hybrid KD                 & 100K &   1M                    \\ \hline
 &                           &                           &                           & 59.57  &  62.70                    \\
\checkmark &                           &                           &                           & 60.29   & 63.15                  \\
                          & \checkmark &                           &                           & 60.44    &62.88                 \\
\checkmark & \checkmark &                           &                           & 61.67    & 63.56                 \\
\checkmark & \checkmark & \checkmark & \multicolumn{1}{l|}{}     & \multicolumn{1}{c|}{62.28} & \multicolumn{1}{c}{63.99} \\
\checkmark & \checkmark &  & \multicolumn{1}{c|}{\checkmark}     & \multicolumn{1}{c|}{62.35} & \multicolumn{1}{c}{64.36} \\
 \hline
\checkmark & \checkmark & \checkmark & \checkmark & \multicolumn{1}{c|}{64.14} & \multicolumn{1}{c}{64.69} \\ \hline \hline
\end{tabular}
\label{Table_ablation}
\end{table}

\subsection{Comparison Study on Two Public Datasets}
\subsubsection{RFMiD Dataset}
Table~\ref{table_rfmid} summarizes the results for the RFMiD dataset. Different from the in-house datasets, it is surprisingly found that RS achieves the best performance among all competitors, which reveals that naive RS can well benefit the test accuracy on the distribution with small ratios, e.g. $\rho = 80$ for the RFMiD dataset. As mentioned above, LDAM requires more time to train, but it still degrades performance terribly even when more training epochs are given. It is also worth noting that almost all of the performance improvements of the comparative methods come from the gains in the tail categories. The resampling-based approaches almost always sacrifice the performance of the head classes, e.g., RS and CBLoss, while the feature-sharing-based approaches almost maintain the performance of the head classes or have only a small loss, e.g., OLTR and RSKD. Our proposed method also achieves competitive results and outperforms all competitors with naive RS applied. 
\subsubsection{ODIR Dataset}
Table~\ref{table_odir} summarizes the results for the ODIR dataset. Our proposed methods achieve the best performance for off-site testing and the second-best accuracy for on-site testing only after RSKD with only a small performance gap. We also observe that both LDAM and CBLoss fail to detect most of the categories, especially for those of group few. It indicates that effective sample-based methods do not work well for those datasets with limited available training samples and a high imbalance ratio. ASL achieves the third-best performance only after RSKD and our proposed methods. As a counterpart, we tried to replace it with ICS in our proposed framework, but no significant improvements were observed, and the same for the other loss functions. It indicates that ICS has compact features and is easier to plug and play.

\subsection{Ablation Study}

\subsubsection{Components Analysis}
To figure out which component makes our methods performant, we have performed an ablation study, and the results are shown in Table~\ref{Table_ablation}. 
As can be seen from Table~\ref{Table_sota}, we first test our proposed MLMC to have the hierarchical information embedded into the model pre-training. We observe that the overall mAP increases from 59.57\% to 60.29\%. 
The adoption of the ICS strategy can also bring a performance gain to the baseline ERM model, with an improvement of 0.87\%. The result of ICS indicates that this sampling strategy can reduce the risk of oversampling on those samples with label co-occurrence. 
Furthermore, we test the effectiveness of combining the MLMC pre-training and ICS strategy, and the overall mAP reaches 61.67\%. The ``cRT" refers to whether we train the ``baseline-ICS" model in a two-stage manner (e.g., cRT~\cite{kang2019decoupling}), that is, whether we freeze the feature encoders in the second stage. Although cRT works well as an independent trick for training long-tailed datasets in most scenarios, it is optional in our proposed framework. It can be noted that cRT brings obvious improvements for the \textit{Retina-100K} dataset, but marginal for a large dataset, i.e., \textbf{Retina-1M}.
We first train the baseline model with MLMC, but using a regular sampling strategy (e.g., class-balanced sampling). 
Then we fine tune the model with the ICS strategy, and the overall mAP is 62.28\% and 63.99\%. 
Finally, the hybrid KD improves the model the most, with a 1.86\% improvement in the average mAP. The same results can be observed in \textit{Retina-1M}. However, ICS seems to have less performance improvement (only 0.18\%) because it has more imbalanced and severe label co-occurrence problems.

\begin{figure}[t]
	\includegraphics[width=9cm]{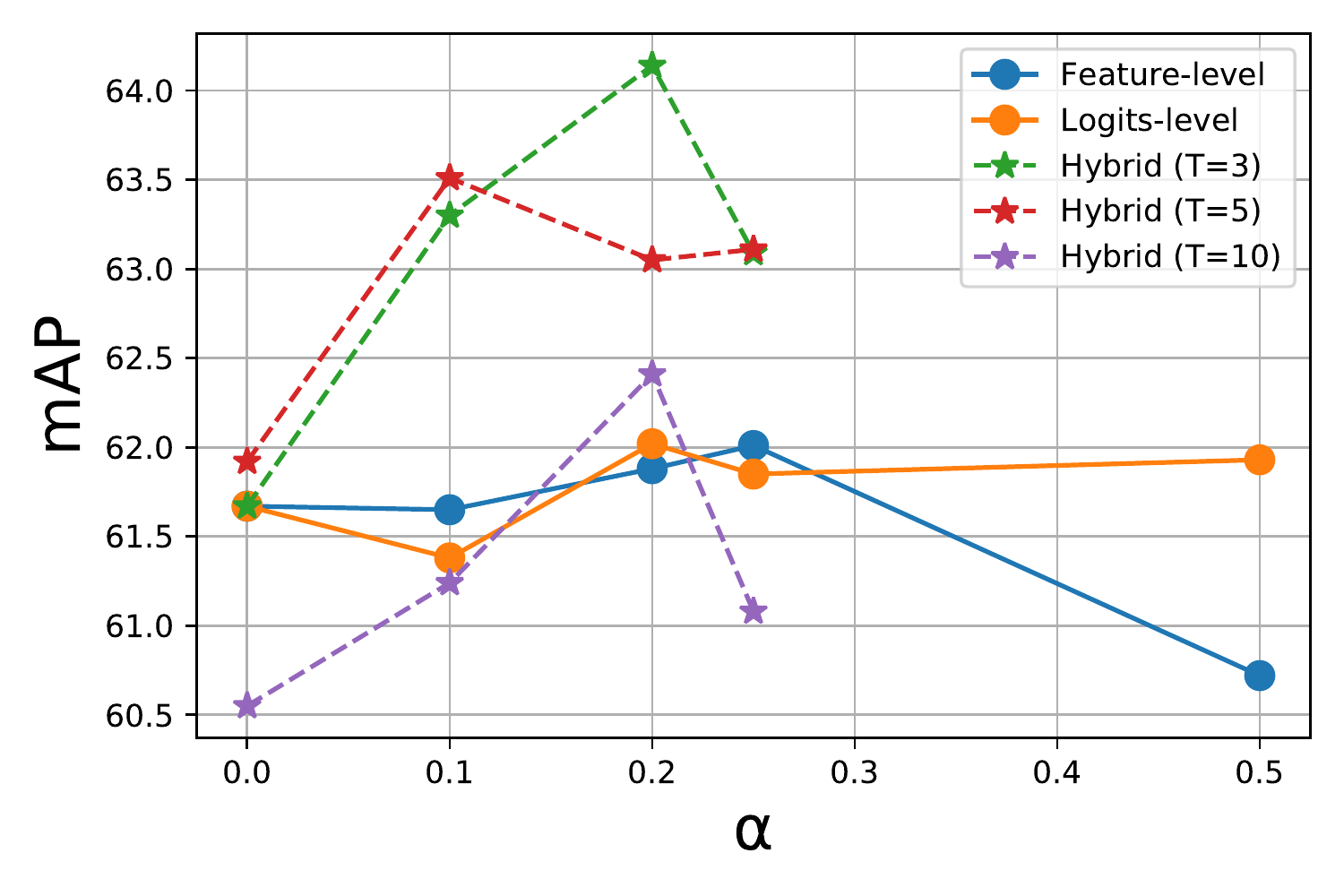}
	\centering
\caption{This figure illustrates the performance variation using different knowledge distillation methods and hyper-parameter value. } \label{fig_kd}
\end{figure}

\subsubsection{Distillation}

In this section, we evaluate different knowledge distillation techniques. We show the overall results on \textit{Retina-100K} in Fig.~\ref{fig_kd} and denote the two basic KD techniques as 'feature-level'~\cite{iscen2021class} and 'logits-level'~\cite{Hinton2015Distilling}, respectively.
Our proposed hybrid KD method consists of two basic KD methods with different intermediate outputs from the DL-based model, such as the features from the last convolutional layer (Eq.~\ref{eq. fkd}) and the logits from the last FC layer (Eq.~\ref{eq. tkd}).  Note that since we keep $\alpha = \beta$ in Eq.~\ref{eq. total_loss}, the range of coefficient selection for $\alpha$ and $\beta$ is between 0 and 0.5. From the results, we can see that both single KD techniques can benefit the baseline model, except when $\alpha = 0.5$ for the feature-level KD, with the 0.95\% loss of accuracy.

Next, we investigate the effect of temperature scaling on model performance. 
Three values are considered and $\mathcal{T} = 3$ in Fig.~\ref{fig_kd} outperforms all counterparts when $\alpha = [0, 0.1, 0.25]$. The best result is obtained with $\mathcal{T} = 3$ and $\alpha = 0.2$. 
However, if we set $\mathcal{T} = 10$, the results do not exceed the baseline model for all $\alpha$ values except $\alpha = 0.2$. It can be concluded that a smaller $\mathcal{T}$ value is beneficial for the KD phase.

\begin{figure*}[t]
    \centering
    \includegraphics[width=18cm]{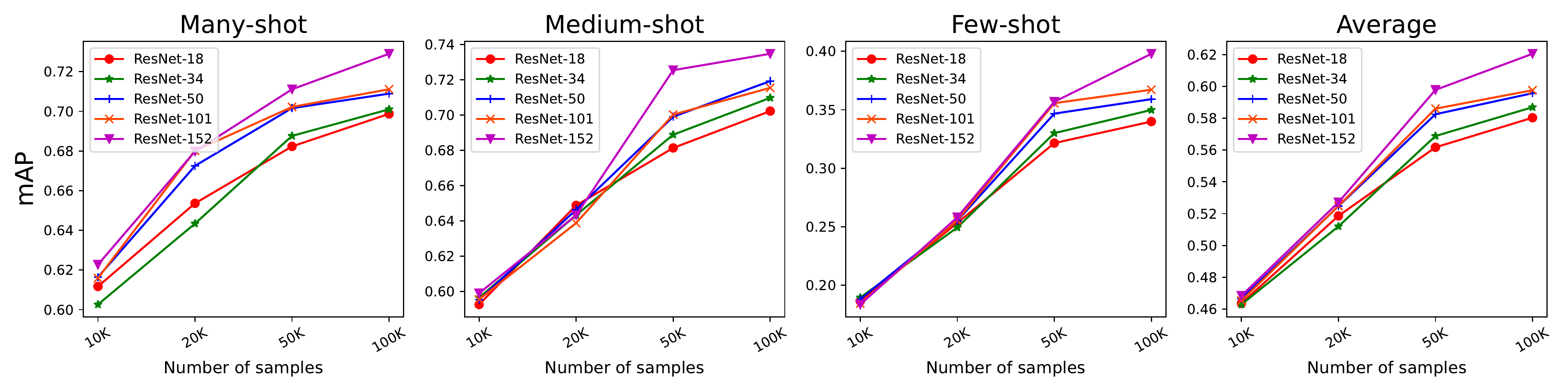}
    \caption{The performance (mAP) changes on different architectures of backbones (\textit{ResNet-18, ResNet-34, ResNet-50, ResNet-101, ResNet-152}) and different scales of datasets (\textit{10K, 20K, 50K, 100K}). } 
    \label{fig_backbone}
\end{figure*}
\subsection{Empirical Analysis on Performance Bound}

In this part, we aim to investigate the upper bound of the retinal diseases recognition model based on some empirical analysis when training from a long-tailed distribution.

\subsubsection{Backbones and Training Samples} First, to investigate how different capacities of a DL-based model would affect the performance, we use the variants of ResNet, which have the same concept but with a different number of layers: \{\textit{ResNet-18, ResNet-34, ResNet-50, ResNet-101, ResNet-152}\}. We then fix the imbalance ratio and sample subsets from \textit{Retina-100K} with different numbers of training samples: \{\textit{10K, 20K, 50K}\}. 
The evaluated results are shown in Fig.~\ref{fig_backbone}. We have the following findings:
\begin{enumerate}
    \item When trained with a small-scale dataset, increasing the capacity of the model has minimum effect on the performance, e.g., from \textit{10K} / \textit{ResNet-18}: 46.36\% mAP to \textit{10K} / \textit{ResNet-152}: 46.84\% mAP with only 0.48\% improvement in terms of average performance. 
    \item In contrast, when using a small capacity version of the model, increasing the training samples only brings marginal performance gain, e.g., for \textit{ResNet-50}, the available training samples are doubled from \textit{50K} to \textit{100K}, but only 1.32\% mAP improvement in terms of average performance.
    \item Rare diseases become easily identifiable when sufficient training samples are provided, e.g., from \textit{100K} / \textit{ResNet-18}: 33.99\% mAP to \textit{100K} / \textit{ResNet-152}: 39.78\% mAP with 5.79\% significant improvement in terms of performance on few-shot classes.
\end{enumerate}

\begin{figure*}[t]
\centering
\subfigure[Baseline Epoch = 5]{
\includegraphics[width=3.5cm]{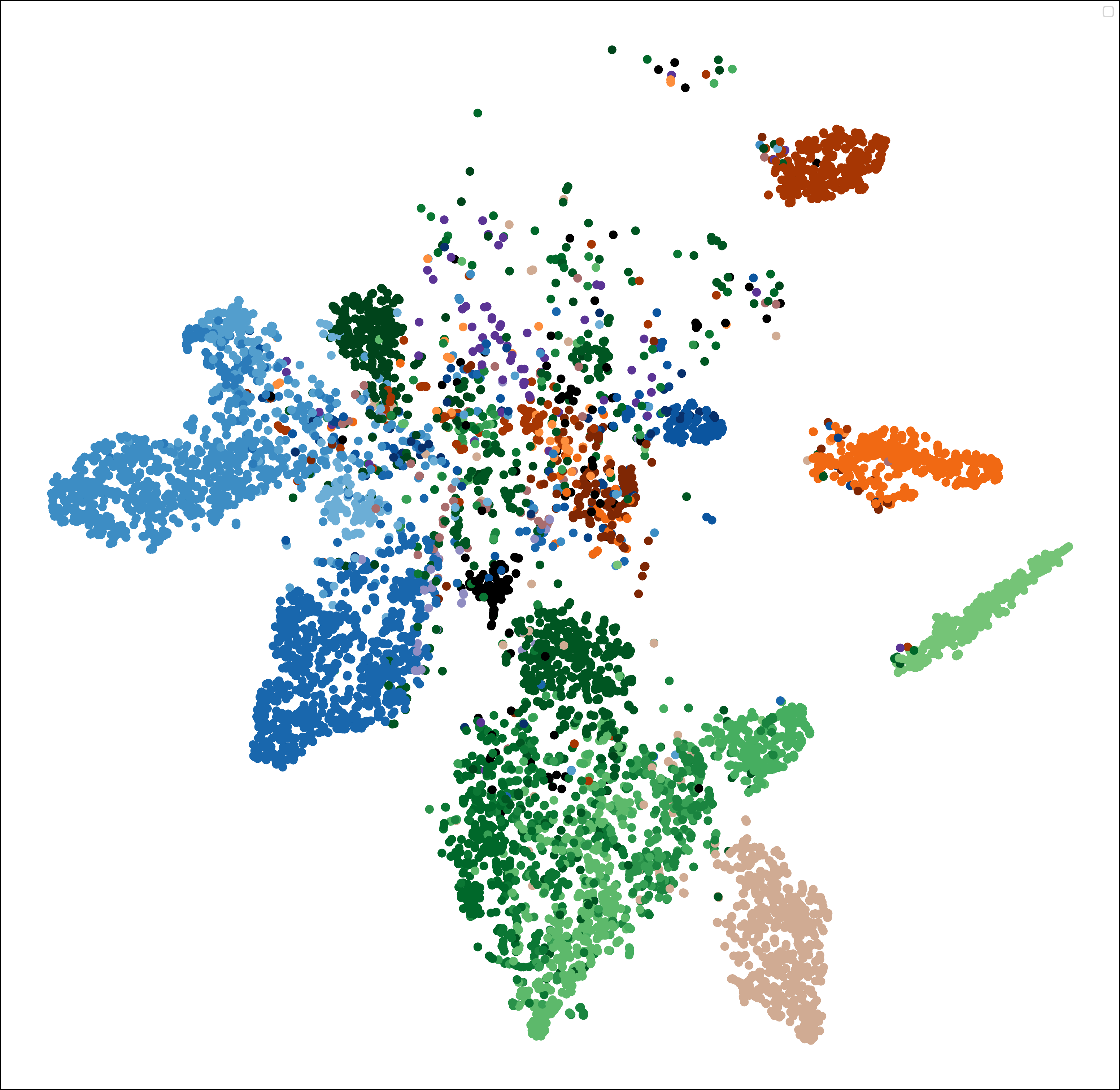}
}
\quad
\subfigure[Baseline Epoch = 45]{
\includegraphics[width=3.5cm]{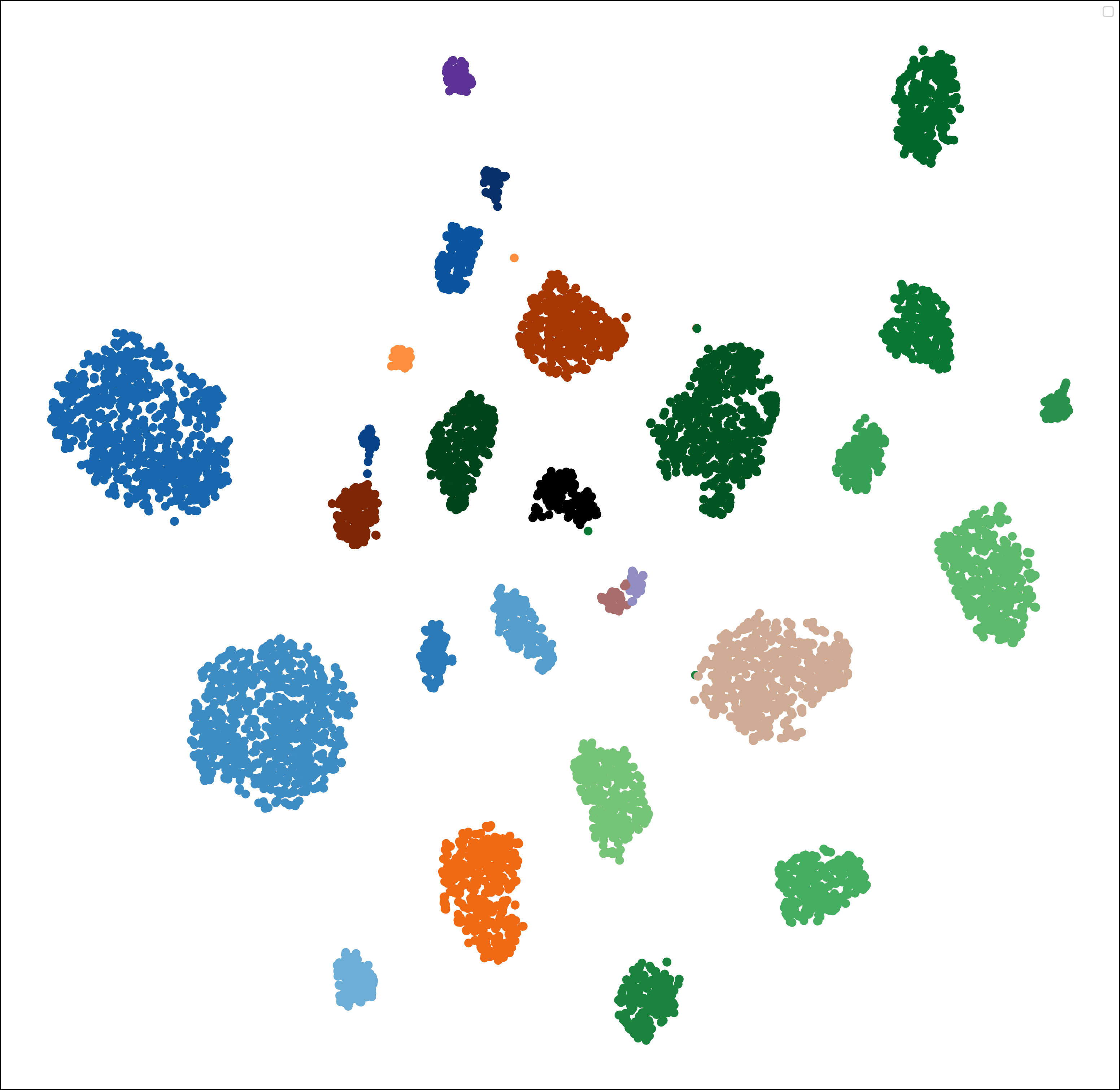}
}
\quad
\subfigure[Ours Epoch = 5]{
\includegraphics[width=3.5cm]{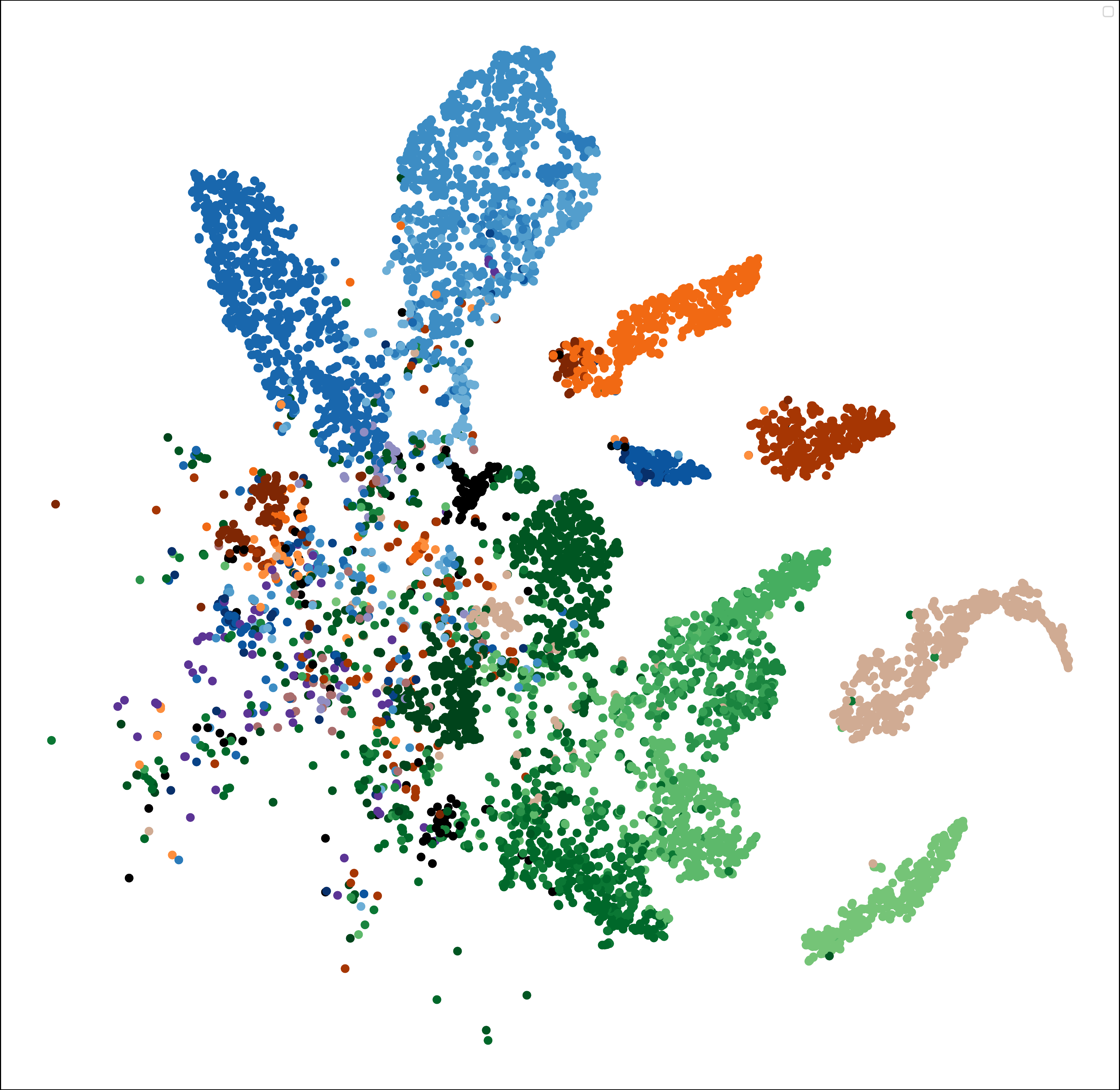}
}
\quad
\subfigure[Ours Epoch = 45]{
\includegraphics[width=3.5cm]{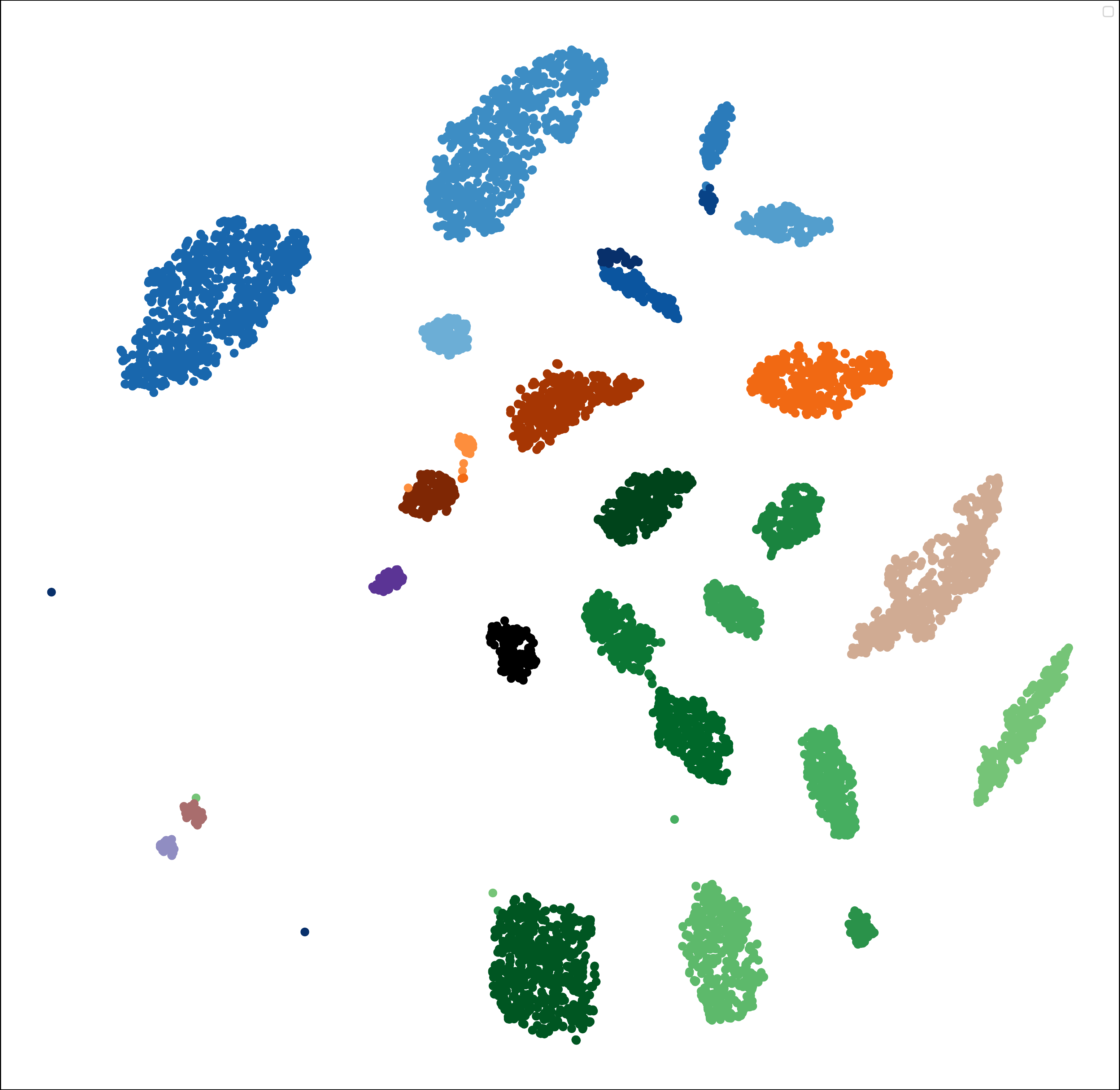}
}
\quad
\subfigure[]{
\includegraphics[width=1.0cm]{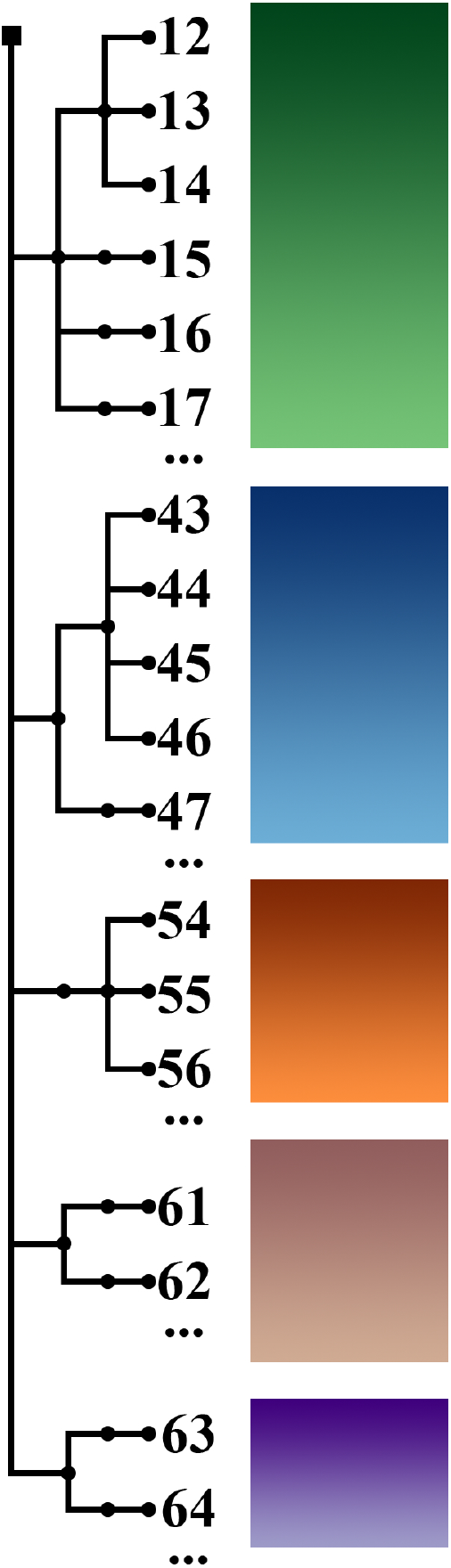}
}

\caption{Learned embedding tSNE visualization. (a) / (b) Results on different epochs of training from BCE Loss without hierarchy-aware constraints. (c) / (d) Results on different epochs of training from BCE Loss with hierarchy-embedded multi-label marginalization classifier (MLMC). (e) The hierarchical tree and the classes that belong to the same parent use similar colors. Clearly, our proposed MLMC method induces closer distance on those classes that belong to the same parent in the feature space while maintaining good discriminations.} 
\label{fig. tsne}
\end{figure*}

\begin{figure*}[t]
    \centering
     \includegraphics[width=18cm]{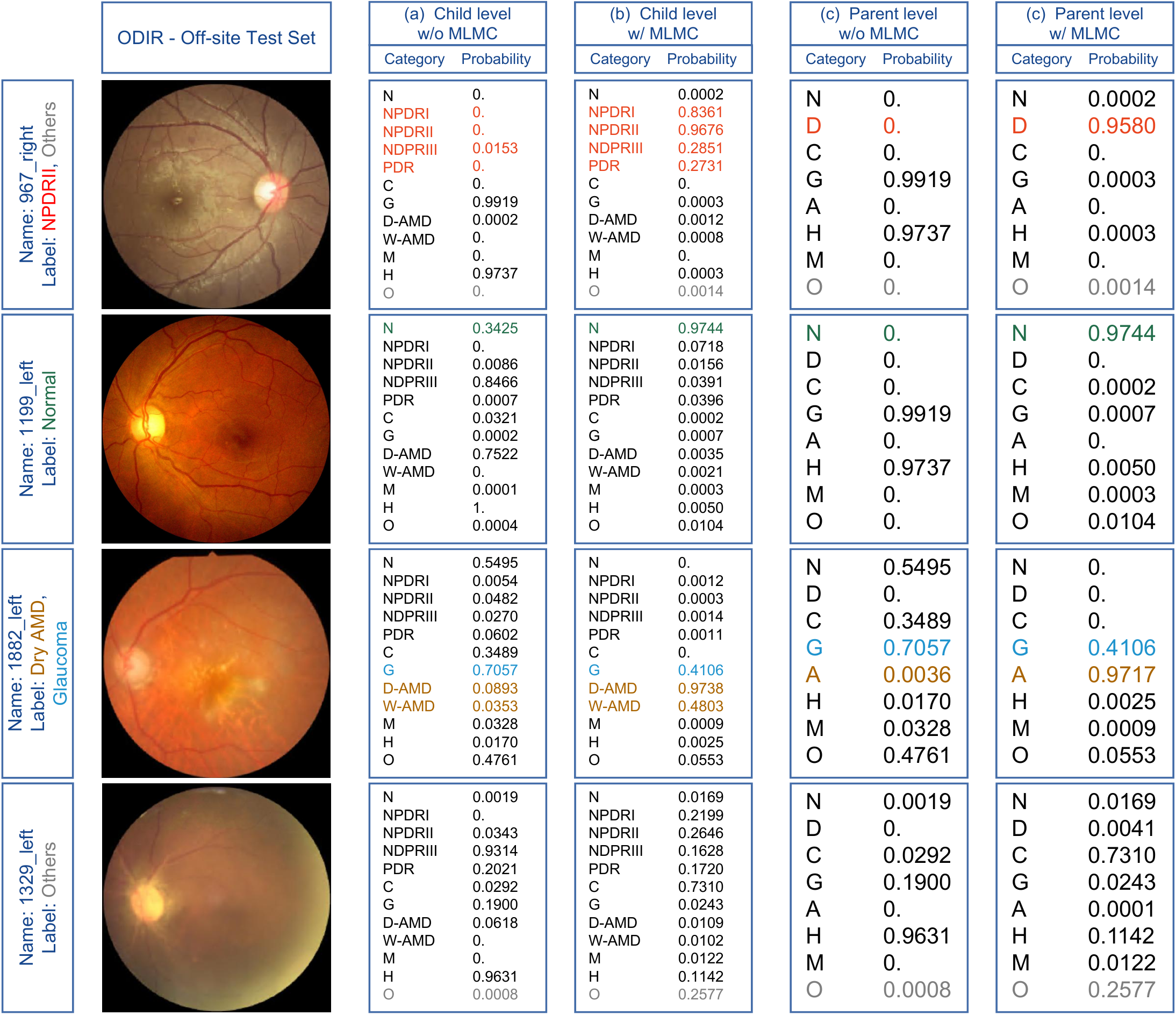}
    \caption{Predicted probabilities from ODIR dataset output by the models w/ or w/o MLMC. The different classes are colored by their corresponding groups. (a) Predicted probabilities from the model w/o MLMC at the child level. (b) Predicted probabilities from the model w/ MLMC at the child level. (c) Predicted probabilities from the model w/o MLMC at the parent level. (d) Predicted probabilities from the model w/ MLMC at the parent level. It can be observed that our proposed methods improve the robustness of the model with more reliable predictions.}
    \label{fig_misdia}
\end{figure*}


\begin{figure*}[t]
    \centering
    \includegraphics[width=18cm]{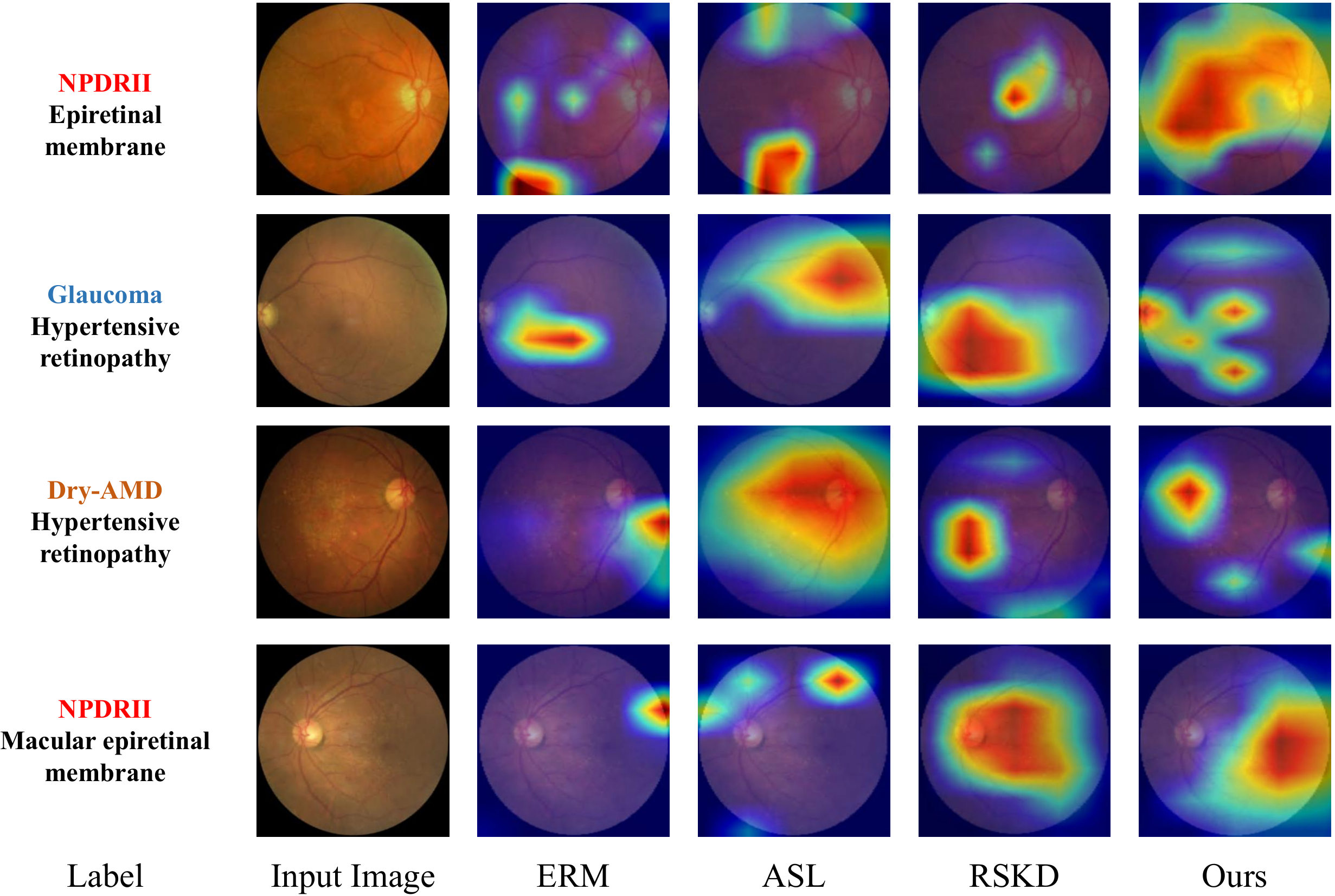}
    \caption{Visualization results of several competitors and our proposed methods. We compare the visualization results with methods w/o hierarchy-aware constraints: ERM and ASL, and methods w/ hierarchy-aware constraints: RSKD and our proposed methods. The original images and corresponding annotations are presented in the first column.} 
    \label{fig_cam}
\end{figure*}

\begin{table}[t]
\small
\centering
\caption{Empirical analysis on the performance bound. A / B denotes training from A and then fine-tuning on B.}
\begin{tabular}{l|ccc|c}
\hline
\hline
           & many   & medium & few    & avg    \\ \hline
10K / 10K   & 61.64 & 59.53 & 18.81 & 46.66 \\
10K / 100K  & 65.74 & 63.67 & 25.94 & 51.78 \\ \hline
20K / 20K   & 67.25 & 64.64 & 25.52 & 52.47 \\
20K / 100K  & 68.25 & 66.40 & 29.17 & 54.61 \\ \hline
50K / 50K   & 70.16 & 69.91 & 34.67 & 58.25 \\
50K  / 100K & 70.58 & 70.99 & 35.22 & 58.93 \\ \hline
100K / 100K & 70.89 & 71.93 & 35.90 & 59.57 \\ \hline \hline
\end{tabular}
\label{table_cRT}
\end{table}
\subsubsection{Decoupling the CNN Components}
Here, we follow the decoupling idea~\cite{kang2019decoupling} to explore the effectiveness of the bottleneck on the long-tailed challenge, which has been posted by \cite{zhang2021distribution}. 
Kang et al.~\cite{kang2019decoupling} proposed to train the representation from the original imbalanced distribution, then freeze the representation layers and fine-tune on the rebalanced distribution. Zhang et al.~\cite{zhang2021distribution} gave a detailed analysis of two-stage methods to investigate the performance bottleneck of two components by retraining the classifier from the original balanced distribution. They concluded that \textit{``training from an imbalanced distribution helps to learn a good representation, and the performance is largely limited by the biased decision boundary of the classifier."} 
Here, we perform experiments with similar settings to investigate the bottleneck of our retinal disease recognition model. 
Since it is difficult to find a balanced distribution due to the co-occurrence of disease labels, we first train the model on a smaller dataset, e.g. \textit{Retina-10K}, then freeze the representation layers and fine-tune the classifier on a larger dataset, e.g. \textit{Retina-100K}. 
The overall results are shown in Table~\ref{table_cRT}. A / B denotes training on A and then fine-tuning on B. 
It can be seen that although the newly introduced training samples do not enrich the feature space (the representation layers are fixed), the classifier can be improved well. 
In particular, 50K / 100K reached a performance closer to the upper bound (100K / 100K), which indicates that \textit{``training from a relatively small dataset can obtain good and rich feature information, but the biased classifier dominates the decision bound, resulting in a huge performance gap."}

\subsection{Qualitative Analysis on Hierarchy-aware Pre-training}
\subsubsection{tSNE visualization}
To better understand how hierarchy-aware, i.e., MLMC constraints for retinal disease detection, we visualize the learned feature representations of the training results w/wo MLMC over tSNE in Fig.~\ref{fig. tsne}. 
It is surprising to find that the features of these similar classes have close distances at an early stage even without any hierarchy-aware constraints, as Fig.~\ref{fig. tsne} - (a) shows. 
This may be due to the fact that the common features of these similar classes are naturally shared. 
However, as the training epochs increase, these similar classes are discriminated by their clear decision boundaries but lose relevance, i.e., larger distances in the feature space, as shown in Fig.~\ref{fig. tsne} - (b). 
Fig.~\ref{fig. tsne} visualizes the learned representation of our proposed methods. It can be observed that the hierarchy-aware constraints can help to induce closer distances on those similar classes while maintaining good discrimination between each other. 
Based on the observations, hierarchy-aware constraints can benefit the learning process across different semantic levels with more efficient sharing of common features among those similar classes.

\subsubsection{Robustness to Shifted Predictions}
In this section, we investigate how hierarchy-aware pre-training helps to improve the robustness of the model, especially against these incorrect predictions from several examples. As Fig.~\ref{fig_misdia} shows, we present the predicted probabilities of two models with and without MLMC at both the child and parent levels. The different classes are colored by their corresponding groups. Note that for the w/o MLMC model, we cannot directly obtain its predictions at the parent level. Here, we directly sum the logits of the child categories with a sigmoid activation function as the predicted probability of the corresponding parent category. 

For the first case, whose ground truth is ``NPDRII, Others", it is incorrectly predicted as glaucoma and hypertensive retinopathy without MLMC. In contrast, the model with MLMC gives correct predictions at both the child and parent levels. Similarly, the second and third cases, whose ground truth is ``Dry AMD, Glaucoma" and ``Normal", are also misdiagnosed as other diseases. MLMC shows robustness to such examples, which are highly correlated, with the additional constraints at the parent level. Note that for the last  case, both models give incorrect predictions where the ground truth is ``Others". This shows a limitation of MLMC, which has difficulty recognizing diseases without obvious hierarchical information or shared features in the semantics. A possible solution is to improve the constraints of semantics and granularity of the hierarchical tree.

Hierarchy-aware pre-training shows great robustness to incorrect predictions. For example, in the first case, the categories with high prediction probability all belong to the same parent category ``DR". In this way, even if the model makes an incorrect prediction at the child level, but with the same parent category, e.g., NPDRII to NPDRI, it carries less risk for practical application compared to predicting as an irrelevant category, e.g., glaucoma. We can think of this as a ``shifted prediction" rather than an ``incorrect prediction". Shifted predictions retain meaningful information for manual reference, and the risk of delayed referral can be reduced.

\subsubsection{Grad-CAM Visualization}
To more directly reflect the effectiveness of hierarchy-aware pre-training, we are interested in what contributes most to the model's predictions. Grad-CAM~\cite{selvaraju2017grad} is a technique that uses the gradients to compute the importance of regions on the images with respect to the final convolutional feature maps. The visualization results are shown in Fig.~\ref{fig_cam}, and all examples selected from the ODIR dataset correctly predict at least one category. For the competitors, we focus mainly on three different approaches: ERM, training without hierarchical constraints but with an extended loss function, e.g. ASL, and training with hierarchical constraints, e.g. RSKD and our proposed methods. From the results, we observe that the ERM-based model can give correct predictions but always presents irrelevant regions. ASL successfully localizes some biomarkers for glaucoma and dry AMD, but still loses the ability to detect some inconspicuous lesions of DR. Compared to non-hierarchical methods, our methods and RSKD pay more attention to those lesions and biomarkers that are crucial for diagnosis. In addition, our proposed methods focus more on the independence between those lesions with different parent categories in a multi-label setting, rather than on large highlighted areas as in RSKD.

\begin{figure}[t]
	\includegraphics[width=8.5cm]{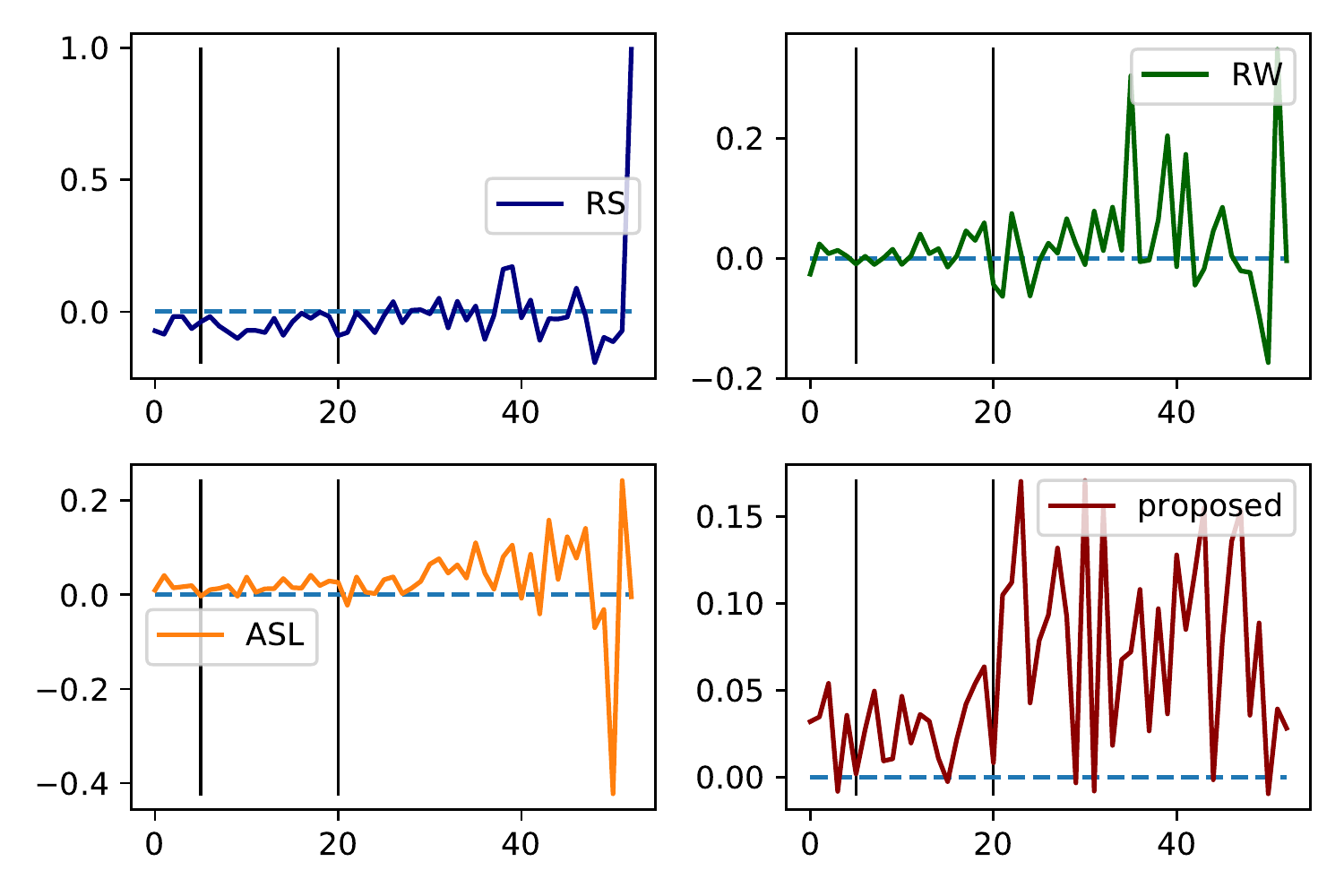}
	\centering
\caption{mAP increments of different classes in groups of shots. The borders between head, medium, and few classes are denoted by the vertical black lines.} \label{fig_increm}
\end{figure}
\subsection{Group-wise Analysis}
\subsubsection{In Groups of Shots}

In Fig.~\ref{fig_increm} we visualize the mAP increments of different classes from different shot-based groups to better see how different approaches affect the performance. 
The classical re-sampling strategy incorrectly samples the label co-occurrence distribution, resulting in a new relatively imbalanced status. 
Several categories in the tail classes obtain improvements in mAP, but the performance of almost all head and middle classes decreases, leading to an overall drop in performance. 
Moreover, we find that RW is agnostic to label co-occurrence, the learned weights of the classifier produce fairer predictions, but the improvements are still marginal. 
ASL and our proposed methods can help the model learn an overall well-generalized representation and a less biased classifier, contributing to the global performance gain.

\subsubsection{In Groups of Diseases}
\begin{table}[t]
\small
\centering
\caption{The comparison of performance on 5 selected parent categories.}
\begin{tabular}{c|ccccc}
\hline \hline
         & Mild  & Macula & Vessels & OD    & Rare  \\ \hline
Original & 66.86 & 55.02  & 41.37   & 32.36 & 18.28 \\
HL~\cite{dhall2020hierarchical} & 67.44 & 57.23 & 45.06 & 33.04 &  17.95 \\
Proposed & \textbf{71.33} & \textbf{58.22}  & \textbf{49.32}   & \textbf{38.63} & \textbf{27.58} \\ \hline \hline
\end{tabular}
\label{table. group_of_dis}
\end{table}

Since we claim that hierarchy knowledge helps to train the model with a well-generalized representation by incorporating more semantic information during the training process, this information can naturally be shared across both coarse and fine categories. 
In Table~\ref{table. group_of_dis}, we show how the proposed methods can improve performance in high/coarse-level classes. 
Here we select five representative coarse diseases, denoted by \textit{mild}, \textit{macula}, \textit{vessels}, \textit{optic disc (OD)}, and \textit{rare} diseases. 
Another widely used training loss in hierarchical classification Hierarchical Loss (HL)~\cite{dhall2020hierarchical} is evaluated as a comparative study. 
We observe that hierarchy-aware pre-training can help improve most categories at a coarse level. However, it cannot solve the problem of data imbalance. 
Our proposed methods can effectively combine hierarchical learning and instance-wise resampling strategy, resulting in performance gain on these rare diseases.

\section{Discussion \& Conclusion}
In this paper, we discuss the necessity and challenges of training a multiple retinal disease recognition model. We propose a novel framework that exploits the hierarchical information as prior knowledge for more efficient training on the feature representations. Moreover, an instance-wise class-balanced sampling strategy and hybrid knowledge distillation are introduced to address the multi-label long-tailed issue. For the first time, we train retinal diseases recognition models from two in-house and two public datasets, one of which includes more than one million fundus images covering more than 50 retinal diseases. The experiment results demonstrate the effectiveness of our proposed methods.

There are some limitations of this work. First, public fundus datasets for long-tailed retinal diseases are still scarce, and our proposed methods show less capacity for those datasets with limited training samples, e.g., ODIR. In addition to collecting more samples, a potential solution is to leverage the knowledge distillation technique to extract rich information from the pre-trained model, to improve the generalization ability of the model for downstream tasks, especially when the private training samples are not available. Second, our proposed methods are not end-to-end and some pre-training phases are needed, which may require more training time and additional adjustment of hyper-parameters towards different scenarios. Third, a human-designed hierarchical tree for retinal diseases is needed for our proposed hierarchy-aware pre-training. To this end, we also released a carefully designed three-level hierarchical tree covering more than 100 kinds of retinal diseases\footnote{Please refer to our supplementary files for more details.}. We hope that it can contribute to the research community in addressing the multi-label long-tailed challenge of retinal disease recognition.

\section{Acknowledgements}
We would like to thank Dr. Danli Shi from Hong Kong Polytechnic University and Haodong Xiao from Xin Hua Hospital Affiliated to Shanghai Jiao Tong University School of Medicine, who helped to organize and design the hierarchical trees used in this work. We would also like to thank Airdoc for the philanthropic funding and the efforts in collecting and annotating the valuable in-house datasets.

\bibliographystyle{IEEEtran}
\bibliography{cite}

\end{document}